\documentclass[10pt,twocolumn,letterpaper]{article}

\usepackage[pagenumbers]{cvpr} 

\usepackage{booktabs,multirow}
\usepackage{algorithm}
\usepackage{algorithmic}
\usepackage{amsmath, amsthm}
\newtheorem{theorem}{Theorem}
\newtheorem{remark}{Remark}

\newcommand{\OurMODEL}{\textsc{D-LEAF}}

\definecolor{cvprblue}{rgb}{0.21,0.49,0.74}
\usepackage[pagebackref,breaklinks,colorlinks,allcolors=cvprblue]{hyperref}

\def\paperID{21371} 
\def\confName{CVPR}
\def\confYear{2026}

\title{Tracing and Mitigating Hallucinations in Multimodal LLMs via Dynamic Attention Localization}

\author{
Tiancheng Yang$^{1,4,*}$,
Lin Zhang$^{2,3,*}$,
Jiaye Lin$^{6}$,
Guimin Hu$^{5}$,
Di Wang$^{2,3}$,
Lijie Hu$^{1,\dagger}$
\\[0.3em]
$^1$MBZUAI\\
$^2$Provable Responsible AI and Data Analytics (PRADA) Lab\\
$^3$King Abdullah University of Science and Technology \\
$^4$School of Advanced Interdisciplinary Sciences, University of Chinese Academy of Sciences \\
$^5$University of Copenhagen\\
$^6$Tsinghua University 
}

\begin{document}
\maketitle
\begin{abstract}
Multimodal Large Language Models (MLLMs) achieve strong performance on tasks like image captioning and visual question answering, but remain prone to hallucinations, where generated text conflicts with the visual input. Prior work links this partly to insufficient visual attention, but existing attention-based detectors and mitigation typically apply uniform adjustments across layers and heads, obscuring where errors originate. In this paper, we first show these methods fail to accurately localize problematic layers. Then, we introduce two diagnostics: Layer Image Attention Entropy (LIAE) which flags anomalous layers, and Image Attention Focus (IAF) which scores attention heads within those layers. Analysis shows that LIAE pinpoints faulty layers and IAF reliably ranks heads that warrant correction. Guided by these signals, we propose Dynamic Layer-wise Entropy and Attention Fusion (D-LEAF), a task-agnostic, attention-guided method that dynamically localizes and corrects errors during inference with negligible overhead. Furthermore, by establishing a connection between D-LEAF and DPO, we provide theoretical justification for the effectiveness of D-LEAF. Results show our D-LEAF delivers a 53\% relative improvement on standard captioning benchmarks, and on VQA both accuracy and F1-score improve by approximately 4\%, substantially suppressing hallucinations while preserving efficiency.
\end{abstract}
\section{Introduction}
Multimodal Large Language Models (MLLMs) have gained increasing attention for their ability to process and integrate visual and textual information. This design allows them to achieve strong performance on a variety of vision-language tasks, such as image captioning, visual question answering, and text-to-image generation \citep{chen2023shikra, zhu2023minigpt, liu2024improved}. However, MLLMs often produce content that contradicts the image or the instructions, which is known as hallucination. These inconsistencies often lead to reliability issues in practical applications, particularly in domains where accuracy and factual consistency are critical \citep{yao2025understanding,he2023geometric, zhang2025modalities,zhou2025flattery,guo2025benchmarking,zhang2024locate,cheng2024leveraging,cheng2024multi,yao2025your,yang2025fraud}.

Traditional strategies to mitigate hallucinations in vision-language models involve instruction fine-tuning or reinforcement learning on carefully curated datasets \citep{gunjal2024detecting, jiang2024hallucination,hu2024dissecting}. Although effective, these approaches are typically resource-intensive and difficult to scale. To overcome these challenges, recent research has shifted to inference‑time methods, mitigating hallucinations by enhancing semantic stability \citep{chen2025mitigating, wang2025tpc, tang2025mitigating,jiang2025msrs,yu2025pixel} or applying contrastive decoding techniques \citep{wang2024mllm, liang2025mole, jiang2025hicd} to adjust the distribution of the final output logits. Although these methods are more effective than training‑based algorithms, they still cannot sufficiently eliminate hallucinations and incur a higher inference latency relative to the baseline. 

In addition, a deeper limitation is mechanistic: prior methods rarely identify \textit{where} hallucinations arise in the attention stack. Several studies implicate over-reliance on the language stream (e.g., ``anchor patterns'' \citep{huang2024opera} or ``textual inertia'' \citep{liu2407paying}) and respond with global adjustments that increase visual weighting \citep{sarkar2025mitigating, jiang2025devils}. However, in practice, we find that these interventions frequently apply undifferentiated suppression across all selected attention modules. This can disrupt correctly functioning heads and thus limit hallucination reduction (see in \cref{motivation}).

\begin{figure}[t]
\centering
\includegraphics[width=\columnwidth]{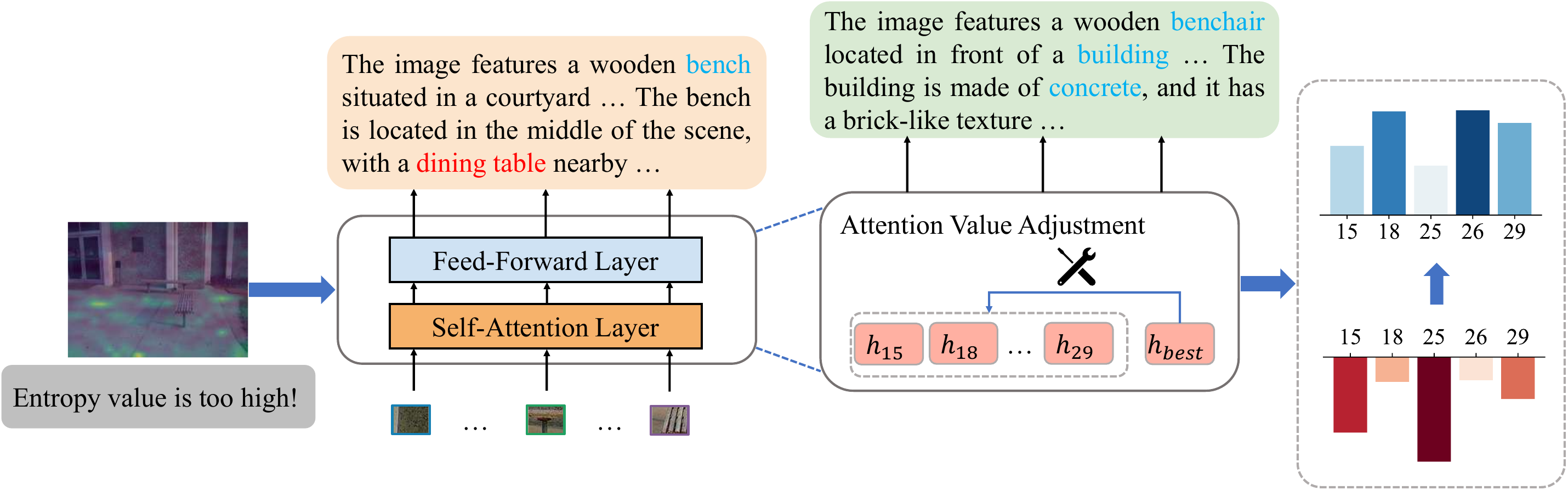}
\vspace{-12pt}
\caption{The workflow of \OurMODEL{}. During inference, when a layer’s attention-module entropy exceeds a dynamic threshold, \OurMODEL{} then corrects the attention heads exhibiting insufficient visual focus, suppressing hallucinations (e.g., the phrase “dining table”).
}
\label{workflow}
\vspace{-12pt}
\end{figure}

To address these issues, we adopt a \textit{localize before correct} strategy. We first introduce two complementary diagnostics that operate during the forward pass: (1) Layer Image-Attention Entropy (LIAE), which flags unreliable layers; and (2) Image-Attention Focus (IAF), which identifies the specific attention heads within those anomalous layers that require correction. Guided by LIAE and IAF, we propose D-LEAF (Dynamic Layer-wise Entropy and Attention Fusion), a lightweight and plug-and-play method that dynamically identifies unreliable attention components and applies selective, fused corrections to only the flagged heads, avoiding blanket suppression. As illustrated in \cref{workflow}, when the attention entropy of a layer exceeds a dynamic threshold, D-LEAF pinpoints low focus heads and injects fused corrective signals, which suppress hallucinated content (e.g., removing the spurious phrase `dining table') while preserving faithful details.

To address these issues, we adopt a \textit{localize before correct} strategy. We first introduce two complementary diagnostics that operate during the forward pass: (1) Layer Image-Attention Entropy (LIAE), which flags unreliable layers; and (2) Image-Attention Focus (IAF), which identifies the specific attention heads within those anomalous layers that require correction. Guided by LIAE and IAF, we propose D-LEAF (Dynamic Layer-wise Entropy and Attention Fusion), a lightweight and plug-and-play method that dynamically identifies unreliable attention components and applies selective, fused corrections to only the flagged heads, avoiding blanket suppression. As illustrated in \cref{workflow}, when the attention entropy of a layer exceeds a dynamic threshold, D-LEAF pinpoints low focus heads and injects fused corrective signals, which suppress hallucinated content (e.g., removing the spurious phrase `dining table') while preserving faithful details.

These results demonstrate that D-LEAF strikes an optimal balance between factual reliability, descriptive detail, and high inference speed. Our contributions are summarized as follows: 
\begin{itemize}
    \item First, we analyze prior attention-head-based suppression methods and show that (i) some attention heads can focus on the correct image information and (ii) blindly suppressing all heads across layers can harm correct ones, leading to ineffective hallucination mitigation. To address this, we propose two inference-time diagnostics LIAE and IAF to dynamically and precisely localize anomalous layers and specific heads requiring correction.
    \item Second, we propose a novel method, called D-LEAF. \OurMODEL{} is a lightweight plug-and-play method that suppresses hallucinations through layer-by-layer corrections during inference: LIAE flags problematic layers, and IAF selects heads to receive fused corrective signals. Furthermore, we show that D-LEAF and DPO both aim to guide the model toward generating preferred answers through a shared optimization objective.
    \item Third, we validate the effectiveness of D-LEAF through extensive experiments on three leading MLLMs in three multimodal hallucination benchmarks, achieving up to a 53\% reduction in hallucinations with only 8\% throughput overhead relative to greedy decoding, without relying on additional tools.
\end{itemize}

\section{Related Work}
\begin{figure}[!htbp]
\centering
\vspace{-12pt}
\includegraphics[width=1.0\columnwidth]{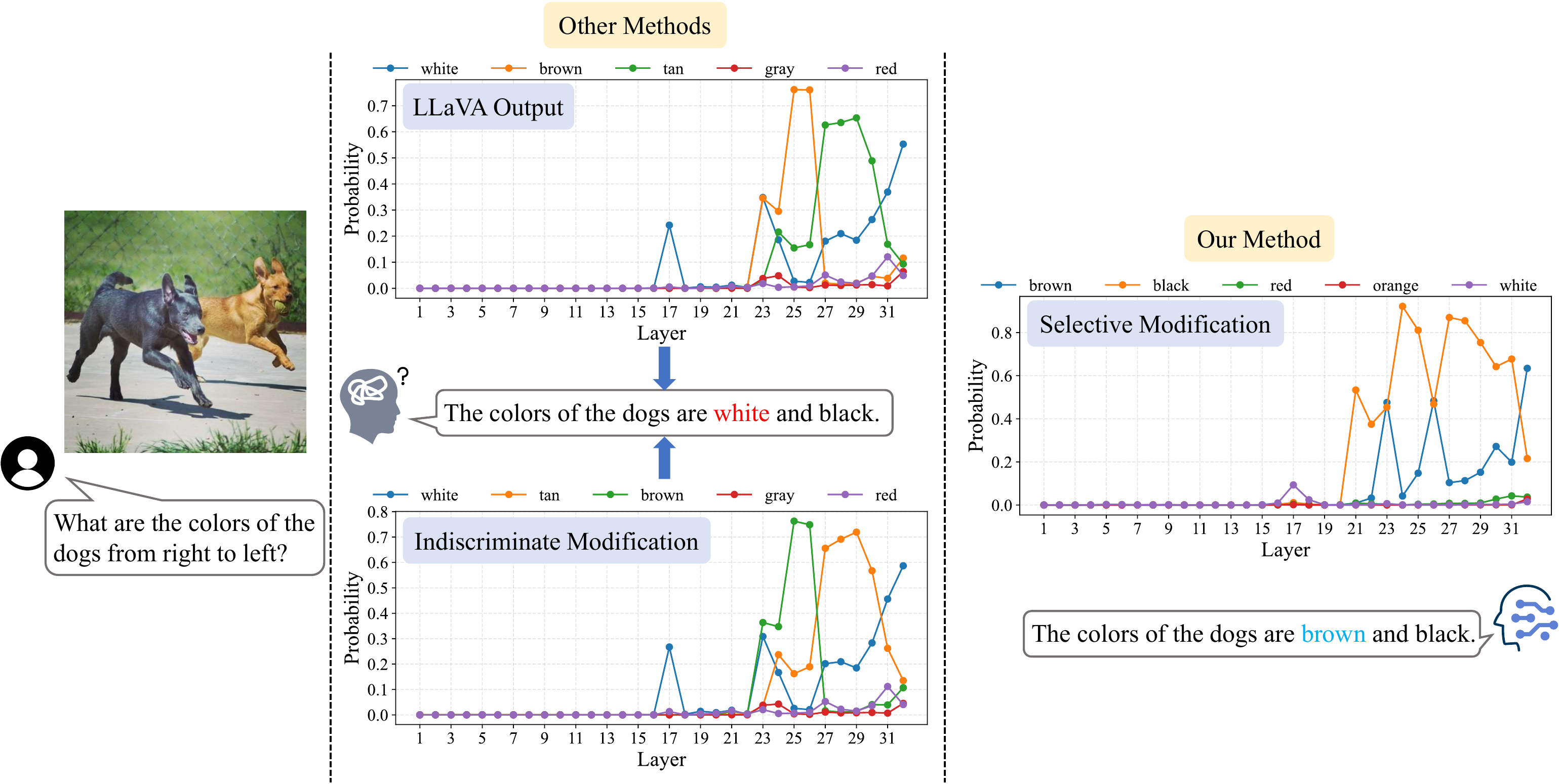}
\caption{A motivating example of using selective attention correction in a visually ambiguous scenario.}
\vspace{-12pt}
\label{motivation}
\end{figure}

\paragraph{Hallucination and Mitigation in MLLMs.}
In natural language processing, hallucinations originally denote generated content that is inconsistent with the context or facts \citep{huang2025survey}. In MLLMs, this manifests as factual errors, incorrect image descriptions, or misidentified object attributes/relationships \citep{liu2024improved}. Previous research on hallucination mitigation can be divided mainly into two categories: training-based algorithms and training-free algorithms. Training-based methods apply visual instruction tuning \citep{gunjal2024detecting}, external expert guidance \citep{chen2024halc}, or reinforcement learning from human feedback (RLHF) \citep{sun2023aligning}, but they typically require substantial compute and are difficult to deploy in resource-constrained settings. Thus, lightweight training-free methods have attracted growing interest. 
A prominent line is contrastive decoding, which mitigates spurious output by comparing model predictions under varying conditions. For example, VCD \citep{leng2024mitigating} contrasts the output distributions conditioned on original and distorted visual inputs to identify and suppress hallucinated content; MoLE \citep{liang2025mole} employs a Mixture of Experts for inter-layer contrast decoding; DAMRO \citep{gong2024damro} reduces the impact of background outlier tokens; OPERA \citep{huang2024opera} performs multiple rollbacks combined with token aggregation to suppress hallucinations; DoLA \citep{chuang2023dola} leverages layer-wise contrasts to enhance factuality; and HALC \citep{chen2024halc} contrasts output distributions across different visual contexts and uses visual matching scores to guide beam-search candidate selection. Despite their effectiveness, these approaches still introduce additional decoding overhead, e.g., HALC incurs a 2.4\text{$\times$} increase in inference time compared to standard greedy decoding. Thus, this motivates us to design a lightweight, plug-and-play method without relying on additional tools.
\paragraph{Interpretability-driven Mitigation in MLLMs Hallucination.}
Numerous studies have examined the underlying causes of hallucinations to guide the development of more fine-grained architecture-level suppression methods. Reported factors include excessive prior knowledge of LLM \citep{liu2407paying}, insufficient attention to images \citep{jiang2025devils, sarkar2025mitigating,you2025mitigating}, and excessive attention to summary words \citep{huang2024opera}. Among these mitigation strategies, attention-head-based hallucination suppression methods show promise. For example, ASCD \citep{wang2025ascd} employs positive and negative steering as two complementary mechanisms to adapt the internal attention distributions of the model. AD‑HH \citep{yangunderstanding} first identifies the heads prone to hallucination offline and then detects and suppresses these heads in real time during the model’s forward pass. In contrast, SPIN \citep{sarkar2025mitigating} and SVAR \citep{jiang2025devils} indiscriminately mute a subset of heads in specific layers to force the model to focus more on visual input. However, because these approaches apply uniform corrections across all layers, they lack flexibility and can still fail to eliminate hallucinations in certain cases, as illustrated in \cref{motivation}. To address these issues, we conducted a systematic analysis of attention‐module behavior during the forward pass of the model and introduced LIAE for layer‐wise detection of problematic heads within each decoder module. By applying targeted per-layer corrections to these specific attention heads, our method more precisely suppresses hallucinations.

\section{Understanding MLLM Hallucination}
In this section, we present empirical analyses to investigate the internal mechanisms behind hallucination in MLLMs from the perspective of attention. Although prior attention-based hallucination mitigation methods have achieved promising results, we observe that they still fail to produce correct answers in semantically ambiguous scenarios, as illustrated in \cref{motivation}. We therefore re-examine these approaches in detail. We hypothesize that \textit{indiscriminately suppression of attention heads across all layers reduces hallucinations but biases the model toward generating shorter outputs.} We aim to answer these three research questions: (i) Are poorly performing heads uniformly distributed across layers?; (ii) How does head suppression reduce hallucinations?; (iii) What costs does suppression of poorly performing heads incur?  

\subsection{Indiscriminate Correction Leads to Errors}

\textbf{Are poorly performing heads uniformly distributed across layers?} Prior attention-correction strategies such as SPIN \citep{sarkar2025mitigating} and SVAR \citep{liu2407paying} suppress the lowest-scoring heads at the intra‐layer level. We argue that ignoring inter-layer head performance can inadvertently suppress functionally correct heads, and thus fail to eliminate hallucinations. As illustrated in the middle of \cref{motivation}, applying SPIN to LLaVA in a visually ambiguous scenario does not prevent the model’s hallucinated outputs. We hypothesize that this phenomenon arises because poorly performing attention heads are not uniformly distributed across all layers, but instead cluster within specific layers.

\begin{figure}[!htbp] 
  \vspace{-13pt}
  \centering
  \includegraphics[width=0.6\linewidth]{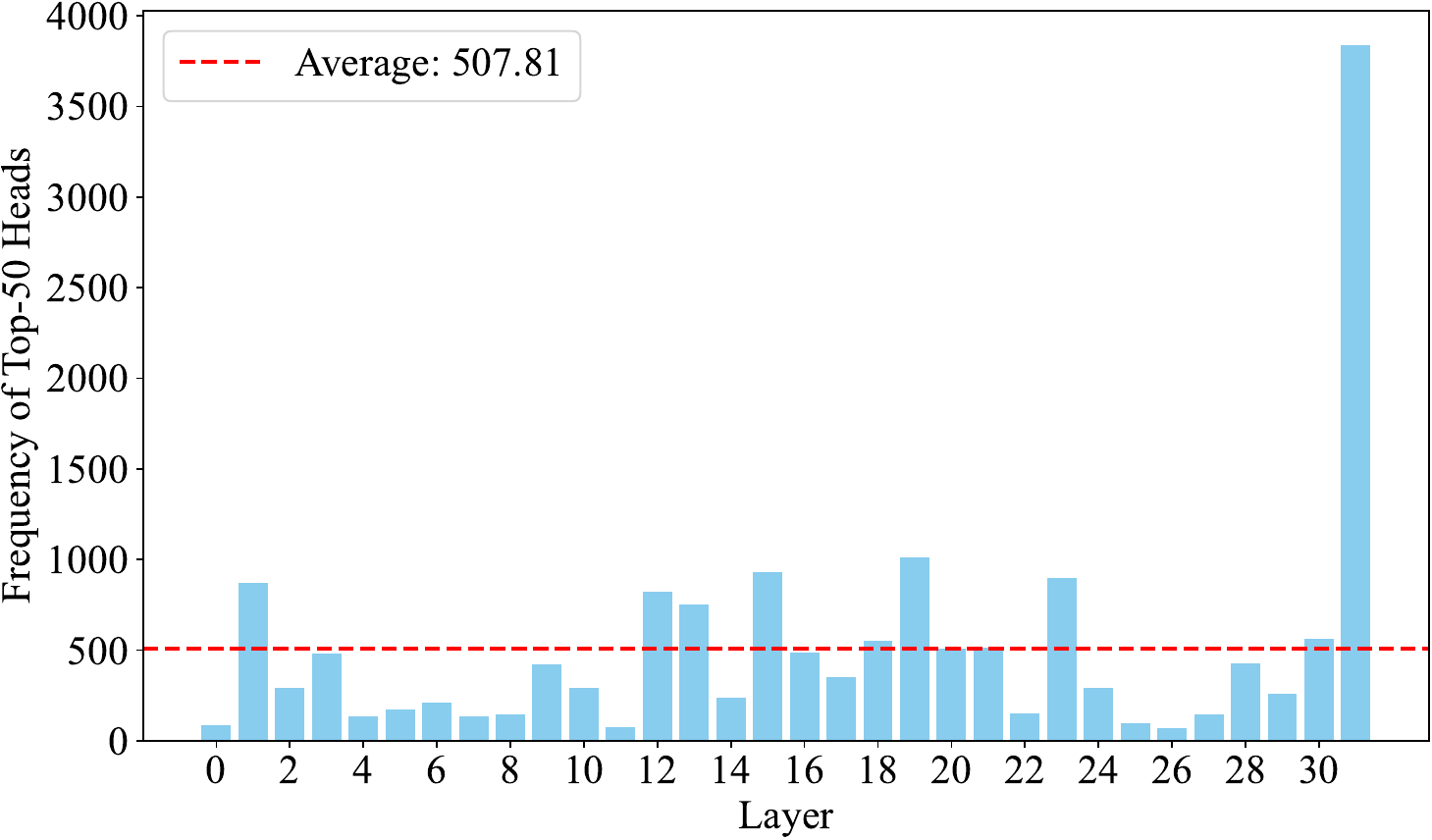}
  \caption{Distribution of abnormal attention heads across layers.}
  \label{abnorm_layer}
  \vspace{-13pt}
\end{figure}

To further verify this hypothesis, we randomly sampled 500 images from the COCO2014 validation set and extracted the hallucinated tokens generated by the model. We rank every attention head across all layers by their SPIN scores, select the 50 lowest performing heads, and visualize their distribution to reveal how these underperformers cluster across layers. As illustrated in \cref{abnorm_layer}, this distribution markedly deviates from uniformity, with the majority of layers falling below the mean. This suggests that indiscriminately modifying the lowest-scoring anomalous attention heads across all layers is not a principled or effective strategy.

Based on the above assumptions, in the context of \cref{motivation}, we rank all attention heads globally by their SPIN scores, select the $k$ worst performing heads (using the same $k$ as SPIN) and suppress them. As shown on the right side of \cref{motivation}, this global suppression successfully prevents the model from hallucination. These results motivate us to design a metric to dynamically localize anomalous attention heads during the model’s forward pass.

\subsection{Muting low-focus heads reduces hallucinations at the cost of accuracy}
\begin{figure}[!htbp] 
  \vspace{-13pt}
  \centering
  \includegraphics[width=0.7\linewidth]{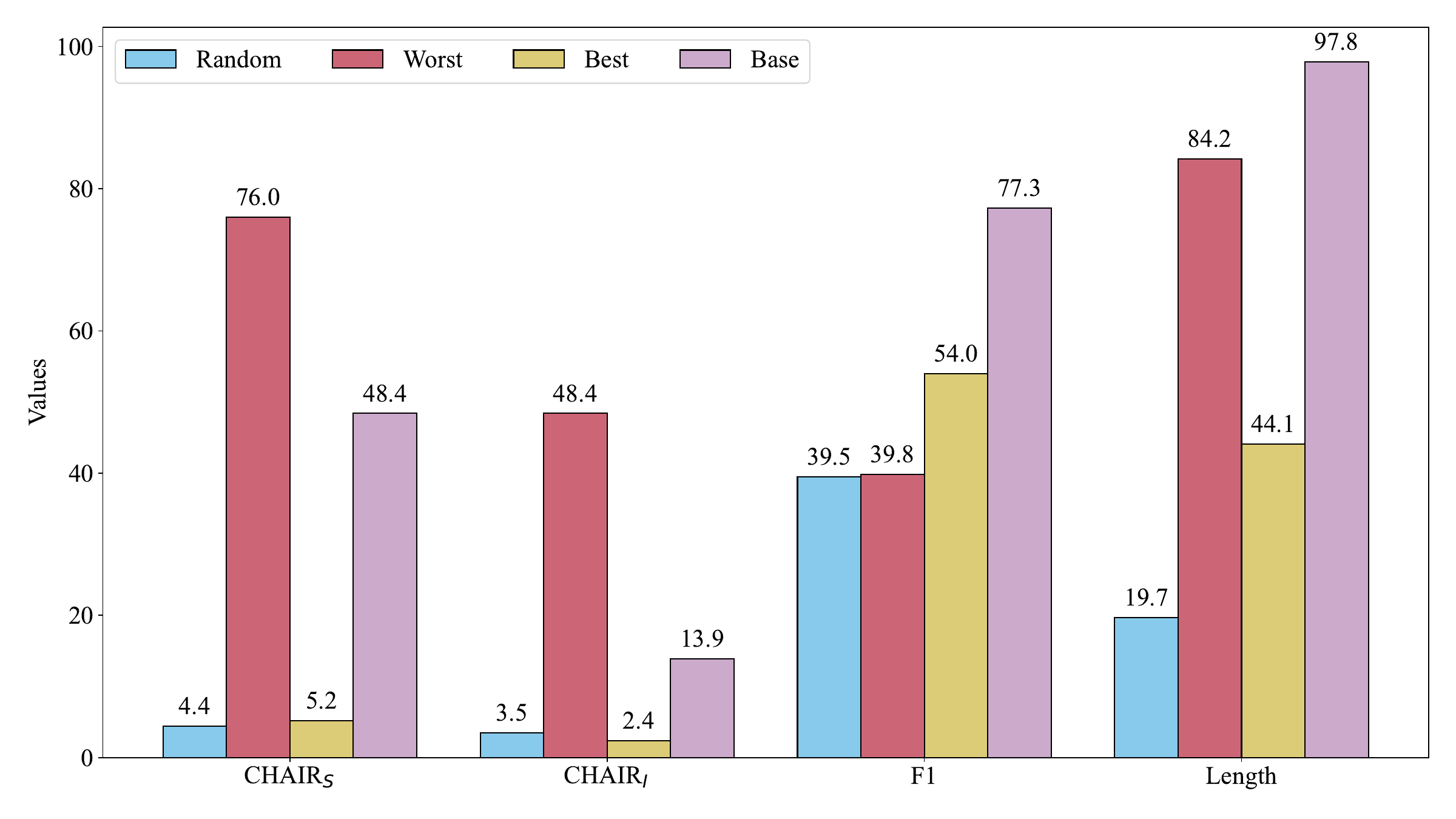}
  \caption{The impact of different suppression methods in LLaVA.}
  \label{llava_supress}
  \vspace{-13pt}
\end{figure}

In this part, we answer the questions `How does head suppression reduce hallucinations?' and `What costs does suppression of poorly performing heads incur?'. Inspired by \citep{li2023inference}, we investigate whether there exist attention heads that can correctly capture image content. Following \citep{rohrbach2018object}, we adopt the CHAIR metrics, namely CHAIR$_I$ and CHAIR$_S$, as defined in \cref{chairi} and \cref{chairs}, and conduct experiments on LLaVA-7B. The detailed definitions of these metrics are provided in the \cref{app:dataset}.

We first partition the attention heads by ranking them within each layer according to their cumulative attention over the image tokens. Based on this ranking, we suppress 15\% of the heads per layer, yielding four experimental settings: (i) no intervention, (ii) suppressing the heads with the highest attention scores, (iii) suppressing the heads with the lowest attention scores, and (iv) randomly suppressing heads. We then evaluate each setting using four metrics: CHAIR$_S$, CHAIR$_I$, F1, and output length (Length), as illustrated in \cref{llava_supress}.

Our results demonstrate that attention heads with high focus on image tokens indeed contribute to visual understanding, as suppressing them leads to a substantial increase in hallucination rates, approximately twice that of the baseline. Conversely, suppressing heads with low image attention can significantly reduce hallucination rates, but this comes at the cost of shorter outputs and a decrease in F1 scores. We attribute this drawback to the indiscriminate suppression of heads across all layers, as discussed above. Moreover, our experiments reveal that modifying only a small subset of attention heads can substantially alter the model’s output, which is consistent with the findings in \citep{kang2025see, kang2025your}.

Building on these observations, we hypothesize that increasing the visual focus of underperforming attention heads can reduce hallucinations, and that self-correction can be achieved by directly leveraging higher-scoring heads within the abnormal layers.

\section{D-LEAF}
\label{dleaf}
We investigate in depth the internal mechanisms behind hallucinations in the last section. In this section, we introduce the Dynamic Layer-wise Entropy and Attention Fusion (D-LEAF) framework to mitigate hallucinations. Our method dynamically detects anomalous behaviors in the MHA modules of MLLMs during the forward pass and applies real-time corrections to the identified problematic components, thereby improving the reliability of the model’s outputs, as shown in \cref{alg}. We begin by introducing a novel metric Layer Image Attention Entropy (LIAE) for detecting anomalous behavior within each decoder module and describe how, once a module is flagged, we use a additional indicator Image Attention Focus (IAF) to pinpoint the exact attention heads that need to be corrected. We also present significance tests and correlation analyses for these metrics (see \cref{app:preliminary} for introductions to these tools). Finally, we detail the complete algorithmic workflow. We have verified the validity and effectiveness of these indicators in \cref{app:metric_val} and replicated all analyses presented in this section on Shikra to validate the generalizability of our proposed metrics; details are provided in \cref{app:genral}.

\subsection{Dynamic Layer Selection}

As discussed before, prior methods typically rank attention heads within each layer and directly suppress those with the lowest scores. However, this intra‑layer ranking ignores cross‑layer context: if a given layer already exhibits higher overall attention scores than other layers, its comparatively weaker heads may still be performing adequately. As a result, suppressing them indiscriminately can fail to reduce and may even exacerbate hallucinations.

\begin{figure}[!htbp] 
  \vspace{-13pt}
  \centering
  \includegraphics[width=0.6\linewidth]{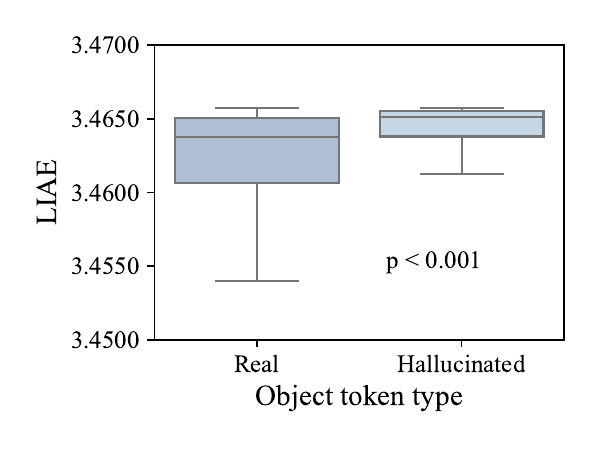}
  \caption{LIAE Distribution across object token types in MiniGPT-4.}
  \label{fig:minigp4_liae}
  \vspace{-13pt}
\end{figure}

Although several layer selection methods have been proposed, particularly in contrastive decoding, for instance, DeCo \citep{wang2024mllm}, which mixes the maximum-probability logits from a selected layer with those of the final layer, and MoLE \citep{liang2025mole}, which adopts a hybrid selection strategy by comparing intermediate logits against the final layer, these approaches rely on a baseline, typically the output of the last layer. However, such a baseline is infeasible for attention-head-based corrections, since modifications to attention occur during the forward pass and each layer’s perturbation directly propagates to subsequent computations. This motivates the need for a new dynamic metric to detect abnormal layers.

Prior work has shown that, during inference in MLLMs, only a subset of attention heads are functionally engaged, and that elevated attention-head entropy is associated with a higher likelihood of hallucination \citep{kang2025your, jiang2025devils}. Therefore, to capture the overall state of all attention heads in the current layer, we first introduce the Maximum Attention Matrix (MAM).  For the $i$-th token of an MLLM output $t_i$, we define the MAM of the $l$-th layer:
\begin{equation}
\label{eq3}
    \text{MAM}_{n}^{(l)} = \max_{h=1,\dots,H} \text{A}^{(l)}_{h,n}, \quad n = 1,\dots,N.
\end{equation}
where $A^{(l)}_{h,n}$ represents the attention of the $h$-th attention head in the $l$-th layer to the image token $n$. Each entry of the MAM at layer $l$ represents, for a given image token $n$, the highest attention score that any of that layer’s attention heads assigns to $n$.

With MAM, we introduce a metric called Layer Image Attention Entropy (LIAE), which quantifies whether a given layer contains attention heads exhibiting overly diffuse focus and therefore require correction.
\begin{equation}
\label{liae}
    \text{LIAE}^{(l)} = - \sum_{n=1}^{N} \text{P}(\text{MAM}_n^{(l)})\text{logP}(\text{MAM}_n^{(l)})
\end{equation}

To validate the effectiveness of the metrics we proposed, we first randomly selected a subset of 500 images from the COCO 2014 validation set \citep{lin2014microsoft}. We chose MiniGPT-4 for subsequent analysis. To show whether LIAE can significantly distinguish the differences between layers when the model generates real words and hallucinated words, we use greedy search in the decoding process of the above model to generate captions for the selected images, prompted by “Please help me describe the image in detail.” We use the ground truth annotation to identify the real and hallucinated words. We then calculated and plotted the distribution of LIAE when the model generated hallucinated words and real words.

To evaluate the significance of these metrics, which independent and non‑normally distributed across real and hallucinated tokens, we apply the Wilcoxon signed‑rank test \citep{wilcoxon1992individual}. With $p < 0.001$, in \cref{fig:minigp4_liae}, we confidently observe in these two models that hallucinated tokens exhibit significantly higher LIAE compared to real tokens.

Accordingly, based on the above experiments, we use LIAE to localize layers that contain anomalous attention heads. For completeness, we report in \cref{app:metric_val} additional experiments that use Layer Image Attention Focus (LIAF) and a hybrid of the two (LIAS) as alternative criteria for abnormal-layer detection, together with ablations. We find that LIAE is more sensitive and achieves the best performance; therefore, we adopt LIAE as the sole metric for abnormal-layer localization.

\subsection{Attention Head Localization}
After pinpointing abnormal layers, we must identify the specific attention heads within them that require modification. Motivated by the “text inertia” phenomenon \citep{liu2407paying}, we introduce Image Attention Focus (IAF), a metric that quantifies the extent to which each attention head attends to image tokens (i.e., visual regions).

\begin{equation}\label{IAF}
\text{IAF}_h^{(l)} = \sum_{n=1}^{N} \text{A}_{h, n}^{(l)}
\end{equation}

\begin{figure}[!htbp] 
  \vspace{-13pt}
  \centering
  \includegraphics[width=0.6\linewidth]{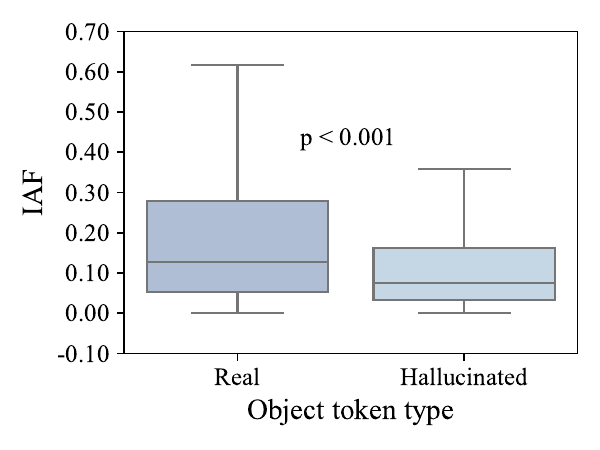}
  \caption{IAF Distribution across object token types in the MiniGPT-4.}
  \label{fig:minigp4_iaf}
  \vspace{-13pt}
\end{figure}
Similarly, to validate the effectiveness of these two metrics, we repeated the Wilcoxon signed‐rank test described above in MiniGPT-4. As shown in \cref{fig:minigp4_iaf}, we plotted the distributions and found that, at p $<$ 0.001, every attention head exhibits significantly higher IAF for real tokens than for hallucinated ones.

Analogous to LIAE, to validate the soundness of our metric, we also evaluate Image Attention Entropy (IAE) and a hybrid of IAF and IAE for identifying and correcting anomalous attention heads; implementation details are provided in \cref{app:metric_val}. Across multiple model architectures, however, we find that using IAF alone consistently achieves the strongest hallucination suppression.

We also verified the reliability of our metrics on other model architectures, with specific results provided in \cref{app:genral}.

\subsection{Mixed Attention Matrix Correction}
Building on the aforementioned layer selection and head localization, we could detect anomalous heads during the model’s forward pass. Our prior experiments indicate that disrupting attention heads with stronger focus on visual tokens causes disproportionately greater degradation and, in particular, increases the model’s propensity to produce hallucinated outputs. Moreover, directly suppressing attention heads with insufficient attention leads to a decline in the quality of the model's output content. Therefore, we hypothesis that hallucinations stem in part from certain heads allocating insufficient attention to visual regions. To remedy this, we introduce a mixed correction approach that leverages well‑performing heads to refine the under‑performing heads.

\begin{equation}\label{mix}
A_{h,v}^l = \gamma A_{best, v}^l + (1 - \gamma) A_{h, v}^l
\end{equation}
where $A_{h,v}^l$ and $A_{best,v}^l$ denote, respectively, the visual‐token submatrices of the attention matrices for the heads requiring correction and the highest‐scoring heads at layer $l$.

\subsection{Dynamic Layer-wise Entropy and Attention Fusion}
Based on these findings, we propose a two‑stage detection and correction algorithm to effectively suppress model hallucinations. Unlike traditional attention‑head‑based hallucination suppression methods, we introduce a dynamic detection baseline Best Attention Score (BAS) that monitors the model’s forward pass and pinpoints the specific layer where anomalies arise.

\begin{equation}\label{bas}
\text{BAS} = \min(\text{BAS}, \text{LIAE}^{(l)}), \quad l=1,2,\dots,L.
\end{equation}

BAS denotes the best score of the layer encountered so far in the forward pass, and it is updated incrementally. During inference, if the current layer’s LIAE is lower than BAS, we set BAS equal to LIAE; if LIAE exceeds BAS, we deem the attention heads in this layer to be underperforming and therefore in need of correction. We then rank all heads in the current layer by their IAF values, select the $n$ worst-performing heads, and update their attention matrices according to \cref{mix}.

Our dynamic localization‑and‑correction algorithm not only suppresses hallucinations at a finer granularity within the model’s architectural layers but also leverages inter‑layer relationships to pinpoint anomalies with greater accuracy. The detailed workflow of the algorithm and its pseudo-code is provided in the \cref{app:code}.
\section{Theoretical Analysis of D-LEAF}

Inspired by previous studies \cite{uppaal2024model, yang2025nullu}, in this section we provide a theorem to analyze the connection between D-LEAF and DPO from the perspective of optimization objectives. This analysis establishes the connection between D-LEAF and DPO, and a more detailed discussion is provided in \cref{appcntdpo}.

Our subsequent analysis is derived by examining a simple logistic model $\pi_{\text{W}}$ for the output conditioned on the input $x$. The conditional probability can be expressed as:
\begin{equation}
\label{logistic}
    \pi_{\text{W}}(y|x_i) = \text{Z}_{\text{W}}^{-1}\text{exp}(o_y^{\top}\text{W}x_i)
\end{equation}
where $w_y$ is the output decoding vector for any $y \in \text{Vocabulary}$, $\text{Z}_\text{W}$ is the normalization factor.

\begin{theorem}[Equivalence of Optimization Directions in DPO and D-LEAF]
\label{thm:dpo_dleaf}
Let $x_i^+$ and $x_i^-$ denote the hallucinated and preferred embeddings at any layer of an MLLM, respectively.  
The gradient of the DPO objective satisfies
\[
\nabla_{\!W} L_{\mathrm{DPO}}
\propto 
\sum_{i=1}^{N} \left( o_{y_i^+}(x_i^+ - x_i^-)^\top 
      + (o_{y_i^+}-o_{y_i^-})(x_i^-)^\top \right).
\]
Thus, DPO reduces the feature difference $x_i^+ - x_i^-$ by pushing the hallucinated embedding toward the preferred one.

In D-LEAF, attention-head fusion modifies the attention matrix so that the updated embedding
\[
\tilde{x}^l 
= \mathrm{FFN}(\mathrm{MHA}_\gamma(x^{l-1}))
\]
decreases the same feature discrepancy 
$\| \tilde{x}^{l} - x_i^- \|$ through attention reweighting.

Therefore, both DPO and D-LEAF optimize toward the same descent direction by reducing the embedding gap between the hallucinated and preferred outputs.
\end{theorem}

\begin{remark}
Theorem~\ref{thm:dpo_dleaf} reveals that D-LEAF is not merely a heuristic attention-correction mechanism. Instead, its update rule can be interpreted as an implicit reparameterization of the DPO gradient in embedding space. Specifically, while DPO explicitly enforces preference alignment by reducing the feature discrepancy $x_i^+ - x_i^-$ through gradient descent, D-LEAF achieves an equivalent effect by modulating the attention matrices that determine each layer's embedding. Consequently, D-LEAF performs a
training-free approximation of the DPO optimization objective, which explains why D-LEAF effectively suppresses hallucinations despite incurring no additional fine-tuning cost.
\end{remark}

\section{Experiment}
\subsection{Experimental Settings}

\paragraph{Datasets.}We conducted a detailed evaluation of our proposed algorithm using three established benchmarks: CHAIR \cite{rohrbach2018object}, POPE \cite{li2023evaluating}, and MMHal‑Bench \cite{sun2023aligning}. These benchmarks were used to assess the effectiveness of our method in suppressing model hallucinations. Detailed descriptions of the benchmarks are provided in \cref{app:dataset}.

\paragraph{Models and Baselines.}
We evaluate our method D-LEAF on 5 models: LLaVA-1.5 (7B) \citep{liu2024improved}, MiniGPT-4 \citep{zhu2023minigpt}, Shikra \citep{chen2023shikra}, InstructBLIP \citep{dai2023instructblip} and Qwen-VL \cite{Qwen-VL}.
And we compare it with several existing SOTA training-free hallucination suppression algorithms, including Greedy Search and Nuclear Sampling, SPIN \citep{sarkar2025mitigating}, PAI \citep{liu2407paying}, VCD \citep{leng2024mitigating}, and DAMRO \citep{gong2024damro}. More details are provided in \cref{app:baseline}.

\paragraph{Evaluation Metric.} CHAIR provided two indicators: C$_I$, which indicates the hallucination rate at the instance-level, and C$_S$, which the hallucination rate at the sentence-level. They are calculated with the following equation: 
\begin{equation}\label{chairi}
\text{CHAIR}_I = \frac{|\{\text{hallucinated objects}\}|}{\text{all mentioned objects}} \end{equation}
\begin{equation}\label{chairs}
\text{CHAIR}_S = \frac{|\{\text{captions with hallucinated objects}\}|}{\text{all captions}}
\end{equation}

\paragraph{Implementation Details.}

All experiments are run in Pytorch using vGPU-48GB. We used a batch size of 1 and set the model’s maximum output length to 512 tokens. For each configuration, we report the average and standard deviation over 3 runs with different random seeds \{42, 927, 111\} in POPE and 5 runs with different random seeds \{42, 3, 11, 927, 111\} in CHAIR. For the image captioning and VQA tasks, we set the number of attention heads that need to be modified on each layer between 3 and 5, $\gamma$ is set within the range of 0.7 to 0.9.

\begin{table*}[t]
  \centering
  \caption{CHAIR hallucination evaluation results. The best result is highlighted in bold, and the second-best is underlined. The values reported are the mean performance.}
  \label{tab2}
  \resizebox{1.0\textwidth}{!}{%
  \begin{tabular}{lcccccccccccc}
    \toprule
    \multirow{2}{*}{Method} 
       & \multicolumn{2}{c}{LLaVA} 
       & \multicolumn{2}{c}{MiniGPT-4} 
       & \multicolumn{2}{c}{Shikra} 
       & \multicolumn{2}{c}{InstructBLIP} 
       & \multicolumn{2}{c}{Qwen-VL} \\
    \cmidrule(r){2-3} \cmidrule(r){4-5} \cmidrule(r){6-7} \cmidrule(r){8-9} \cmidrule(r){10-11}
       & C$_S$ & C$_I$ & C$_S$ & C$_I$ & C$_S$ & C$_I$ & C$_S$ & C$_I$ & C$_S$ & C$_I$ \\
    \midrule
    Greedy   & 47.08 $_{\scriptsize \pm 1.54}$ & 13.00 $_{\scriptsize \pm 0.59}$ 
             & 34.00 $_{\scriptsize \pm 1.98}$ & 10.82 $_{\scriptsize \pm 0.59}$ 
             & 54.64 $_{\scriptsize \pm 2.84}$ & 14.96 $_{\scriptsize \pm 1.37}$ 
             & 48.12 $_{\scriptsize \pm 2.98}$ & 14.18 $_{\scriptsize \pm 1.12}$ 
             & 46.88 $_{\scriptsize \pm 1.56}$ & 12.72 $_{\scriptsize \pm 0.36}$ \\
    Sampling & 53.44 $_{\scriptsize \pm 2.21}$ & 16.30 $_{\scriptsize \pm 1.26}$ 
             & 33.80 $_{\scriptsize \pm 1.17}$ & 11.78 $_{\scriptsize \pm 0.62}$ 
             & 57.10 $_{\scriptsize \pm 1.80}$ & 16.14 $_{\scriptsize \pm 0.88}$ 
             & 47.04 $_{\scriptsize \pm 1.47}$ & 13.24 $_{\scriptsize \pm 0.77}$ 
             & 47.04 $_{\scriptsize \pm 1.47}$ & 13.24 $_{\scriptsize \pm 0.77}$ \\
    VCD      & 55.38 $_{\scriptsize \pm 1.17}$ & 15.20 $_{\scriptsize \pm 1.21}$ 
             & -- & -- 
             & 55.16 $_{\scriptsize \pm 2.25}$ & 14.96 $_{\scriptsize \pm 1.26}$ 
             & -- & -- 
             & 51.40 $_{\scriptsize \pm 2.21}$ & 13.62 $_{\scriptsize \pm 0.75}$ \\
    PAI      & 35.28 $_{\scriptsize \pm 1.69}$ &  9.46 $_{\scriptsize \pm 0.67}$  
             & 27.92 $_{\scriptsize \pm 1.47}$ & 10.06 $_{\scriptsize \pm 0.89}$ 
             & 54.64 $_{\scriptsize \pm 2.51}$ & 14.02 $_{\scriptsize \pm 1.49}$ 
             & 59.24 $_{\scriptsize \pm 2.00}$ & 16.10 $_{\scriptsize \pm 0.67}$ 
             & 47.64 $_{\scriptsize \pm 2.26}$ & 12.92 $_{\scriptsize \pm 0.68}$ \\
    DAMRO    & 46.44 $_{\scriptsize \pm 1.66}$ & 12.78 $_{\scriptsize \pm 0.60}$ 
             & -- & -- 
             & -- & -- 
             & -- & -- 
             & -- & -- \\
    SPIN     & \underline{29.04 $_{\scriptsize \pm 2.46}$} & \underline{8.70 $_{\scriptsize \pm 0.55}$} 
             & \underline{24.56 $_{\scriptsize \pm 1.62}$} & \underline{9.40 $_{\scriptsize \pm 1.74}$} 
             & \underline{38.56 $_{\scriptsize \pm 2.35}$} & \underline{10.88 $_{\scriptsize \pm 0.56}$} 
             & \underline{48.80 $_{\scriptsize \pm 2.65}$} & \underline{14.04 $_{\scriptsize \pm 1.23}$} 
             & \underline{33.72 $_{\scriptsize \pm 2.54}$} & \underline{9.42 $_{\scriptsize \pm 0.48}$} \\
    D-LEAF   & \textbf{23.44 $_{\scriptsize \pm 2.63}$} & \textbf{6.72 $_{\scriptsize \pm 0.49}$} 
             & \textbf{11.56 $_{\scriptsize \pm 1.69}$} & \textbf{4.72 $_{\scriptsize \pm 0.95}$} 
             & \textbf{26.35 $_{\scriptsize \pm 1.32}$} & \textbf{10.62 $_{\scriptsize \pm 0.87}$} 
             & \textbf{22.44 $_{\scriptsize \pm 2.75}$} & \textbf{8.48 $_{\scriptsize \pm 5.93}$} 
             & \textbf{25.24 $_{\scriptsize \pm 1.55}$} & \textbf{7.96 $_{\scriptsize \pm 1.65}$} \\
    \bottomrule
  \end{tabular}
  }
\end{table*}

\begin{table*}[t]
  \centering
  \caption{Quantitative comparison on Muti-turn POPE. The best result is highlighted in bold, and the second-best is underlined. The values reported are the mean performance. }
  \label{tab3}
  \resizebox{\textwidth}{!}{%
  \begin{tabular}{ll*{3}{cc}}
    \toprule
    \multirow{2}{*}{Model}        & \multirow{2}{*}{Method}  & \multicolumn{2}{c}{Random} & \multicolumn{2}{c}{Popular} & \multicolumn{2}{c}{Adversarial}\\
    \cmidrule{3-4} \cmidrule{5-6} \cmidrule{7-8}
    & & Accuracy & F1 & Accuracy & F1 & Accuracy & F1 \\
    \midrule
    \multirow{6}{*}{LLaVA} 
                 & Greedy   & 86.63 $_{\pm 0.78}$  & 85.32 $_{\pm 0.92}$  & 79.22 $_{\pm 0.23}$  & 78.23 $_{\pm 0.66}$  & 76.98 $_{\pm 0.16}$ & 76.71 $_{\pm 0.41}$\\
                 & Sampling & 83.94 $_{\pm 0.59}$  & 83.30 $_{\pm 0.44}$  & 76.30 $_{\pm 0.15}$  & 75.05 $_{\pm 0.35}$  & 72.70 $_{\pm 1.61}$ & 73.44 $_{\pm 2.85}$\\
                 & PAI      & 77.02 $_{\pm 4.62}$  & 72.96 $_{\pm 6.94}$  & 75.68 $_{\pm 0.12}$  & 72.65 $_{\pm 0.25}$  & 75.07 $_{\pm 0.55}$  & 72.41 $_{\pm 0.75}$\\
                 & DAMRO    & \underline{86.67 $_{\pm 0.87}$}  & \underline{85.53 $_{\pm 1.09}$}  & 79.23 $_{\pm 0.28}$  & 78.28 $_{\pm 0.71}$  & 77.01 $_{\pm 0.17}$  & 76.74 $_{\pm 0.42}$\\
                 & SPIN     & 86.29 $_{\pm 0.16}$ & 84.80 $_{\pm 0.38}$ & \underline{81.81 $_{\pm 2.85}$} & \underline{80.47 $_{\pm 2.60}$} & \underline{80.31 $_{\pm 4.33}$} & \underline{79.47 $_{\pm 4.01}$}\\
                 & D-LEAF   & \textbf{87.76 $_{\pm 0.47}$}    & \textbf{86.65 $_{\pm 0.72}$}    & \textbf{84.75 $_{\pm 3.21}$}    & \textbf{83.67 $_{\pm 3.00}$}    & \textbf{84.94 $_{\pm 3.35}$}    & \textbf{82.09 $_{\pm 2.87}$}\\
    \midrule
    \multirow{5}{*}{MiniGPT-4}
                 & Greedy   & 70.17 $_{\pm 3.68}$  & 68.33 $_{\pm 3.82}$ & 64.01 $_{\pm 4.39}$ & 63.16 $_{\pm 5.57}$ & 63.06 $_{\pm 2.48}$ & 62.52 $_{\pm 1.17}$\\
                 & Sampling & 70.34 $_{\pm 5.48}$ & 60.96 $_{\pm 9.00}$ & 62.92 $_{\pm 5.22}$ & 54.41 $_{\pm 8.31}$ & 60.30 $_{\pm 4.70}$ & 55.52 $_{\pm 7.78}$\\
                 & PAI      & 68.45 $_{\pm 9.67}$ & 64.12 $_{\pm 7.14}$ & 59.66 $_{\pm 8.49}$ & 57.31 $_{\pm 6.45}$ & 60.93 $_{\pm 6.52}$ & 56.38 $_{\pm 5.42}$\\
                 & SPIN     & \underline{72.89 $_{\pm 2.44}$} & \underline{69.04 $_{\pm 2.19}$} & \underline{66.10 $_{\pm 2.36}$} & \underline{64.33 $_{\pm 4.18}$} & \underline{64.88 $_{\pm 3.05}$} & \textbf{67.34 $_{\pm 2.66}$}\\
                 & D-LEAF   & \textbf{74.95 $_{\pm 0.62}$} & \textbf{72.17 $_{\pm 0.77}$}    & \textbf{67.51 $_{\pm 0.09}$}    & \textbf{66.02 $_{\pm 0.15}$}    & \textbf{67.62 $_{\pm 0.41}$}    & \underline{65.17 $_{\pm 0.46}$}\\
    \midrule
    \multirow{5}{*}{Shikra}
                 & Greedy   & 80.75 $_{\pm 0.56}$ & 80.35 $_{\pm 0.54}$ & 74.95 $_{\pm 1.88}$ & 75.66 $_{\pm 1.22}$ & \underline{73.37 $_{\pm 1.97}$} & \underline{75.89 $_{\pm 1.03}$}\\
                 & Sampling & \underline{81.67 $_{\pm 0.65}$} & \underline{81.55 $_{\pm 0.69}$} & \underline{77.58 $_{\pm 0.36}$} & \underline{78.67 $_{\pm 0.51}$} & 72.94 $_{\pm 1.45}$ & 75.37 $_{\pm 0.68}$\\
                 & PAI      & 71.00 $_{\pm 0.25}$ & 74.03 $_{\pm 0.22}$ & 70.10 $_{\pm 0.39}$ & 73.38 $_{\pm 0.60}$ & 64.97 $_{\pm 0.62}$ & 70.65 $_{\pm 0.09}$\\
                 & SPIN     & 64.68 $_{\pm 0.03}$ & 61.66 $_{\pm 0.78}$ & 59.73 $_{\pm 0.16}$ & 60.06 $_{\pm 0.61}$ & 58.17 $_{\pm 0.11}$ & 61.09 $_{\pm 0.71}$\\
                 & D-LEAF   & \textbf{82.36 $_{\pm 1.12}$} & \textbf{83.32 $_{\pm 1.14}$}    & \textbf{79.12 $_{\pm 0.61}$} & \textbf{79.89 $_{\pm 0.73}$} & \textbf{75.15 $_{\pm 2.10}$}    & \textbf{76.37 $_{\pm 1.45}$}\\
    \midrule
    \multirow{5}{*}{InstructBLIP}
                 & Greedy   & 86.16 $_{\pm 0.65}$ & 84.54 $_{\pm 0.86}$ & 84.48 $_{\pm 1.10}$ & 83.03 $_{\pm 1.28}$ & 81.95 $_{\pm 0.25}$ & 80.75 $_{\pm 0.46}$\\
                 & Sampling & 79.36 $_{\pm 0.20}$ & 78.36 $_{\pm 0.31}$ & 76.55 $_{\pm 0.32}$ & 76.17 $_{\pm 0.18}$ & 74.74 $_{\pm 0.20}$ & 74.93 $_{\pm 0.05}$\\
                 & PAI      & 86.34 $_{\pm 0.44}$ & 84.66 $_{\pm 0.63}$ & 84.80 $_{\pm 0.94}$ & 83.22 $_{\pm 1.08}$ & \textbf{82.69 $_{\pm 0.03}$} & \textbf{81.32 $_{\pm 0.12}$}\\
                 & SPIN     & \underline{86.51 $_{\pm 0.55}$} & \underline{85.06 $_{\pm 0.76}$} & \underline{85.21 $_{\pm 1.07}$} & \underline{83.91 $_{\pm 1.24}$} & 81.97 $_{\pm 0.19}$ & 80.96 $_{\pm 0.42}$\\
                 & D-LEAF   & \textbf{86.67 $_{\pm 0.57}$} & \textbf{85.23 $_{\pm 0.74}$}    & \textbf{85.32 $_{\pm 0.96}$} & \textbf{84.06 $_{\pm 1.05}$} & \underline{82.10 $_{\pm 0.24}$}    & \underline{81.09 $_{\pm 0.42}$}\\
    \midrule
    \multirow{5}{*}{Qwen-VL}
             & Greedy   & \underline{89.56 $_{\pm 0.24}$} & \underline{89.20 $_{\pm 0.28}$} & 86.93 $_{\pm 0.05}$ & 86.76 $_{\pm 0.02}$ & 82.81 $_{\pm 0.11}$ & 83.25 $_{\pm 0.02}$\\
             & Sampling & 85.41 $_{\pm 0.32}$ & 84.82 $_{\pm 0.38}$ & 81.49 $_{\pm 0.25}$ & 81.42 $_{\pm 0.14}$ & 77.82 $_{\pm 1.23}$ &  72.24 $_{\pm 0.62}$\\
             & PAI      & 89.34 $_{\pm 0.11}$ & 88.89 $_{\pm 0.11}$ & \underline{86.94 $_{\pm 0.30}$} & \underline{86.80 $_{\pm 0.22}$} & \underline{82.88 $_{\pm 0.57}$} & \underline{83.32 $_{\pm 0.40}$}\\
             & SPIN     & 88.49 $_{\pm 0.03}$ & 88.22 $_{\pm 0.06}$ & 84.66 $_{\pm 1.13}$ & 84.86 $_{\pm 0.99}$ & 80.49 $_{\pm 0.80}$ & 81.48 $_{\pm 0.62}$\\
             & D-LEAF   & \textbf{89.59 $_{\pm 0.27}$} & \textbf{89.21 $_{\pm 0.22}$}    & \textbf{87.42 $_{\pm 0.13}$} & \textbf{87.18 $_{\pm 0.11}$} & \textbf{83.19 $_{\pm 0.30}$}    & \textbf{83.50 $_{\pm 0.19}$}\\
    \bottomrule
  \end{tabular}
}
\end{table*}

\begin{figure}[t]
\centering
\vspace{-12pt}
\includegraphics[width=0.8\linewidth]{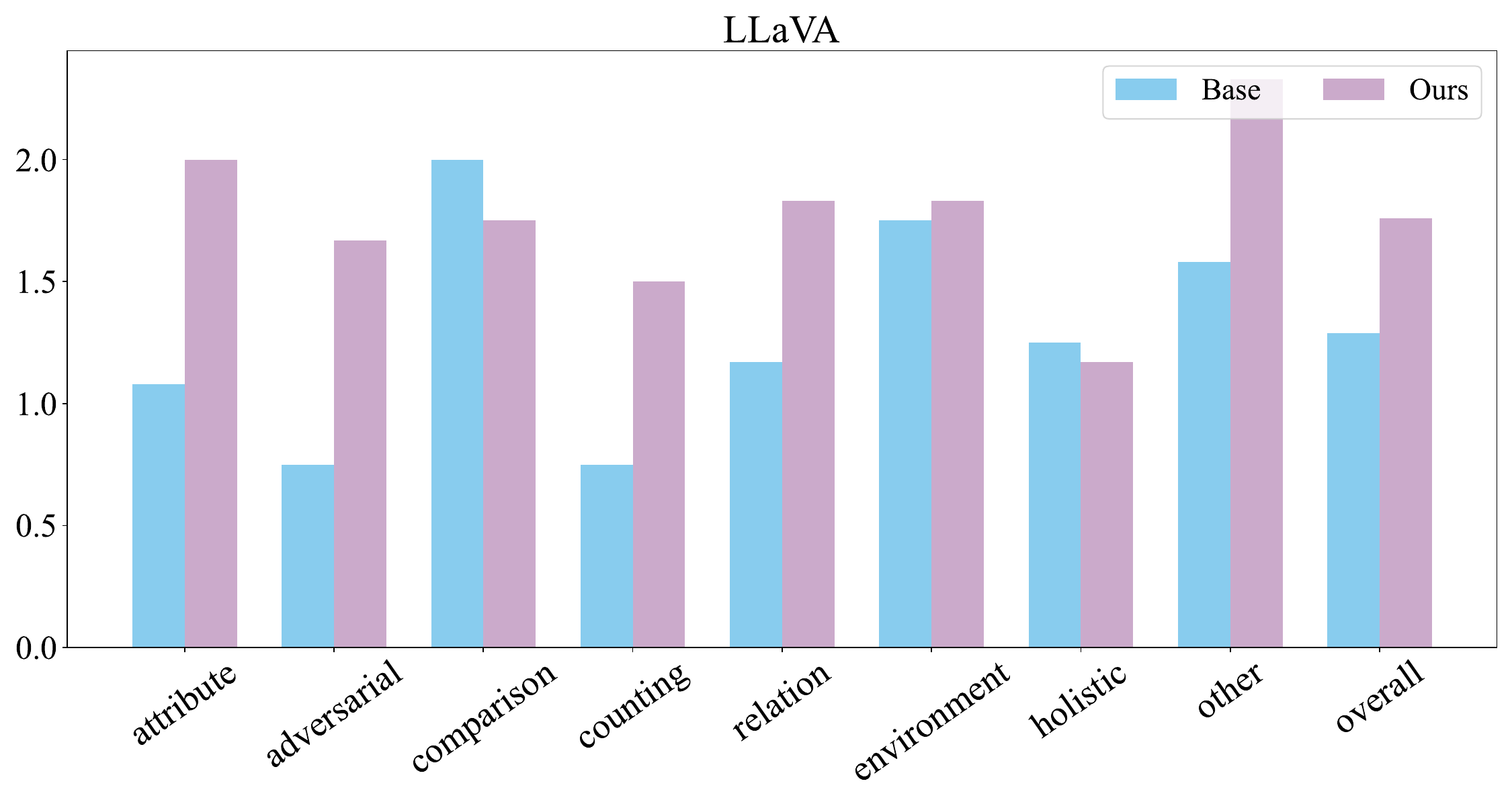}
\caption{MMHal-Bench Evaluation on LLaVA.}
\label{mmbench}
\vspace{-12pt}
\end{figure}

\subsection{Main Results}

\paragraph{Long Sequence Hallucination Evaluation.} We evaluate the CHAIR result of five models, as presented in \cref{tab2}. Our D-LEAF method significantly outperforms previous state-of-the-art approaches across all metrics on hallucination and models. Specifically, on MiniGPT-4 our model achieves a 53$\%$ reduction in $\text{C}_S$ and a 57$\%$ reduction in $\text{C}_I$ compared to SPIN, highlighting the effectiveness of D-LEAF in mitigating object hallucinations in long text generation tasks.

\paragraph{Multi-turn Hallucination Evaluation.} We use a multi-turn POPE evaluation to increase the difficulty of this task, the result is shown in \cref{tab3}. D-LEAF consistently performs best across each part of the POPE across the five models. Notably, on LLaVA, MiniGPT-4 and Shikra, our model outperforms the baseline by approximately 5$\%$, and on the other two models, D-LEAF also surpasses the current state-of-the-art by 1$\%$. The results indicated that our D-LEAF could achieve good results in long context VQA tasks.


\paragraph{GPT-4 Assisted Hallucination Evaluation in Comprehensive General Scenarios.} We use MMHal-Bench and GPT-4 assist to evaluate the performance of D-LEAF in more complex scenarios. From \cref{mmbench}, the experimental results indicate that our method could achieve better results in in LLaVA 7B. However, for comparison-type tasks involving multi-object attribute reasoning, the model’s performance is relatively weak. We also provide the results of MiniGPT-4 and Shikra on MMHal-Bench in \cref{appmmhal}, where our method consistently improves upon the baseline performance on average.

\begin{figure}[t] 
\vspace{-12pt}
\centering
\includegraphics[width=0.6\linewidth]{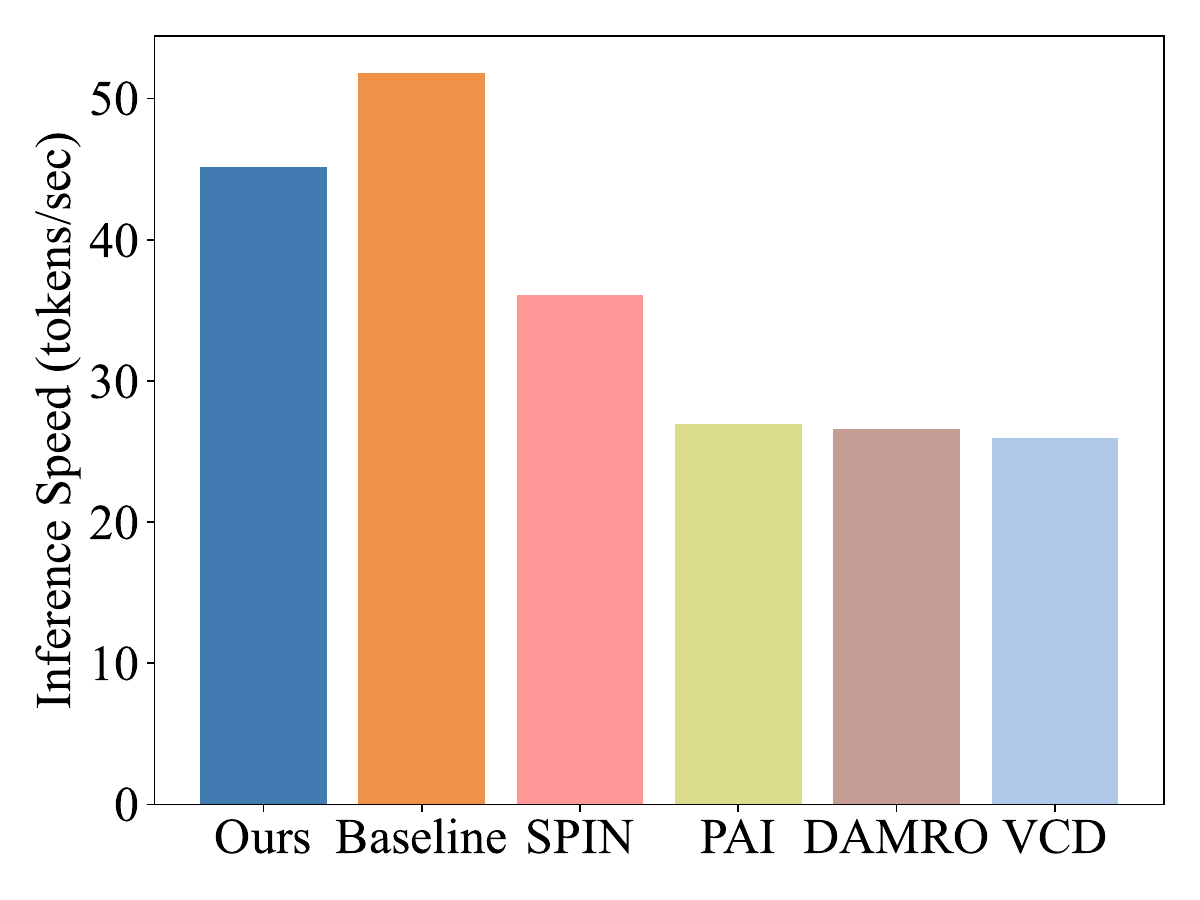}
\caption{Throughput comparison with existing methods in LLaVA.}
\label{fig:llava_speed}
\vspace{-12pt}
\end{figure}

\paragraph{Throughput Estimation.} To evaluate whether our algorithm maintains real-time efficiency without incurring significant throughput loss, we measured the token-per-second generation rate of LLaVA under different algorithms, as shown in \cref{fig:llava_speed}. Our method showed the least reduction in throughput compared to the baseline. It outperformed other state-of-the-art techniques, including attention-head correction methods such as SPIN and PAI. We repeated this experiment on MiniGPT-4 and Shikra and observed consistent results, as detailed in \cref{app:throughput}. 

In \cref{app:case}, we provide visualizations across diverse MLLMs to  further present instances of hallucination corrections by our method.

\subsection{Ablation Study}
\begin{figure}[t]
\centering
\includegraphics[width=1.0\columnwidth]{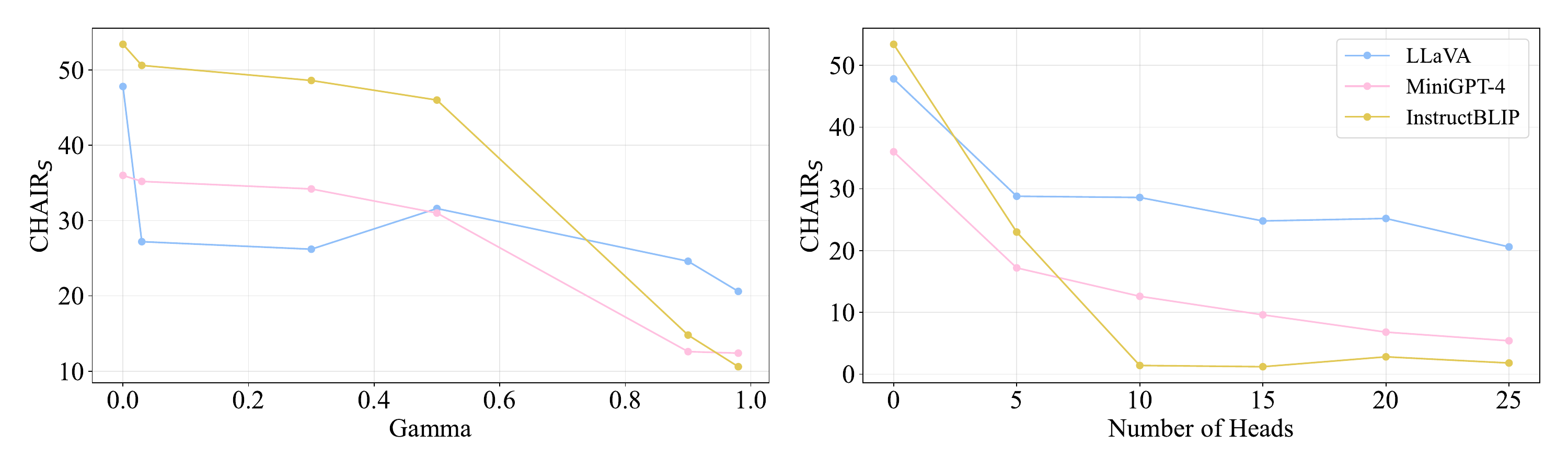}
\caption{Ablation Study results for hyperparamter $\gamma$ and $n$.}
\label{CS_abl}
\vspace{-12pt}
\end{figure}

D-LEAF incorporates two primary hyperparameters: $\gamma$ and the number of heads $n$. We present the effect of varying two parameters on CHAIR$_S$ in the main text, as illustrated in \cref{CS_abl}. The results demonstrate the strong robustness of our method: across a wide range of $\gamma$ (0.03 to 0.98) and $n$ values (5 to 25), our algorithm consistently outperforms the baseline. In the \cref{app:abl}, we further provide the impact of hyperparameter variations on CHAIR$_I$ and F1, along with a more detailed analysis of the results. Moreover, to ensure the completeness of our study, we also examine whether restricting the D-LEAF algorithm to specific layers yields additional gains. The results in \cref{app:abl} and \cref{app:visual} demonstrate that, unlike PAI \cite{liu2407paying} and Deco \cite{wang2024mllm}, D-LEAF consistently achieves significant suppression of hallucinations regardless of whether the layer prior is applied.

\section{Conclusion}

We proposed D-LEAF to suppress the hallucinations generated by MLLMs. We propose a two-stage localization and correction algorithm: first, we use the Layer Image Attention Entropy to identify anomalous modules during the forward pass; then, we apply the Image Attention Focus to rank that layer’s heads and selectively correct the lowest-performing ones. Our experiments demonstrated that D-LEAF outperforms existing methods in reducing hallucinations across various MLLMs.
This work highlights the potential of attention modules to enhance the output reliability of MLLMs and provides mechanistic insights into their operation. 

{
    \small
    \bibliographystyle{ieeenat_fullname}
    \bibliography{main}

@String(AAAI = {AAAI})

@article{cheng2024leveraging,
  title={Leveraging logical rules in knowledge editing: A cherry on the top},
  author={Cheng, Keyuan and Ali, Muhammad Asif and Yang, Shu and Lin, Gang and Zhai, Yuxuan and Fei, Haoyang and Xu, Ke and Yu, Lu and Hu, Lijie and Wang, Di},
  journal={arXiv preprint arXiv:2405.15452},
  year={2024}
}

@article{zhang2024locate,
  title={Locate-then-edit for multi-hop factual recall under knowledge editing},
  author={Zhang, Zhuoran and Li, Yongxiang and Kan, Zijian and Cheng, Keyuan and Hu, Lijie and Wang, Di},
  journal={arXiv preprint arXiv:2410.06331},
  year={2024}
}

@article{guo2025benchmarking,
  title={Benchmarking and Mitigate Psychological Sycophancy in Medical Vision-Language Models},
  author={Guo, Zikun and Xu, Xinyue and Xiang, Pei and Yang, Shu and Han, Xin and Wang, Di and Hu, Lijie},
  journal={arXiv preprint arXiv:2509.21979},
  year={2025}
}

@article{yao2025your,
  title={Is your llm-based multi-agent a reliable real-world planner? exploring fraud detection in travel planning},
  author={Yao, Junchi and Xu, Jianhua and Xin, Tianyu and Wang, Ziyi and Zhu, Shenzhe and Yang, Shu and Wang, Di},
  journal={arXiv preprint arXiv:2505.16557},
  year={2025}
}

@article{yang2025fraud,
  title={Fraud-r1: A multi-round benchmark for assessing the robustness of llm against augmented fraud and phishing inducements},
  author={Yang, Shu and Zhu, Shenzhe and Wu, Zeyu and Wang, Keyu and Yao, Junchi and Wu, Junchao and Hu, Lijie and Li, Mengdi and Wong, Derek F and Wang, Di},
  journal={arXiv preprint arXiv:2502.12904},
  year={2025}
}

@article{hu2024dissecting,
  title={Dissecting representation misalignment in contrastive learning via influence function},
  author={Hu, Lijie and Ren, Chenyang and Xie, Huanyi and Saadi, Khouloud and Yang, Shu and Tan, Zhen and Zhang, Jingfeng and Wang, Di},
  journal={arXiv preprint arXiv:2411.11667},
  year={2024}
}

@article{yu2025pixel,
  title={PIXEL: Adaptive Steering Via Position-wise Injection with eXact Estimated Levels under Subspace Calibration},
  author={Yu, Manjiang and Li, Hongji and Singh, Priyanka and Li, Xue and Wang, Di and Hu, Lijie},
  journal={arXiv preprint arXiv:2510.10205},
  year={2025}
}

@article{jiang2025msrs,
  title={MSRS: Adaptive Multi-Subspace Representation Steering for Attribute Alignment in Large Language Models},
  author={Jiang, Xinyan and Zhang, Lin and Zhang, Jiayi and Yang, Qingsong and Hu, Guimin and Wang, Di and Hu, Lijie},
  journal={arXiv preprint arXiv:2508.10599},
  year={2025}
}

@article{yao2025understanding,
  title={Understanding the repeat curse in large language models from a feature perspective},
  author={Yao, Junchi and Yang, Shu and Xu, Jianhua and Hu, Lijie and Li, Mengdi and Wang, Di},
  journal={arXiv preprint arXiv:2504.14218},
  year={2025}
}

@article{cheng2024multi,
  title={Multi-hop question answering under temporal knowledge editing},
  author={Cheng, Keyuan and Lin, Gang and Fei, Haoyang and Yu, Lu and Ali, Muhammad Asif and Hu, Lijie and Wang, Di and others},
  journal={arXiv preprint arXiv:2404.00492},
  year={2024}
}

@article{zhang2025modalities,
  title={When Modalities Conflict: How Unimodal Reasoning Uncertainty Governs Preference Dynamics in MLLMs},
  author={Zhang, Zhuoran and Wang, Tengyue and Gong, Xilin and Shi, Yang and Wang, Haotian and Wang, Di and Hu, Lijie},
  journal={arXiv preprint arXiv:2511.02243},
  year={2025}
}

@article{zhou2025flattery,
  title={Flattery in Motion: Benchmarking and Analyzing Sycophancy in Video-LLMs},
  author={Zhou, Wenrui and Yang, Shu and Yang, Qingsong and Guo, Zikun and Hu, Lijie and Wang, Di},
  journal={arXiv preprint arXiv:2506.07180},
  year={2025}
}

@article{you2025mitigating,
  title={Mitigating Behavioral Hallucination in Multimodal Large Language Models for Sequential Images},
  author={You, Liangliang and Yao, Junchi and Yang, Shu and Hu, Guimin and Hu, Lijie and Wang, Di},
  journal={arXiv preprint arXiv:2506.07184},
  year={2025}
}

@article{zhu2023minigpt,
  title={Minigpt-4: Enhancing vision-language understanding with advanced large language models},
  author={Zhu, Deyao and Chen, Jun and Shen, Xiaoqian and Li, Xiang and Elhoseiny, Mohamed},
  journal={arXiv preprint arXiv:2304.10592},
  year={2023}
}

@article{chen2023shikra,
  title={Shikra: Unleashing multimodal llm's referential dialogue magic},
  author={Chen, Keqin and Zhang, Zhao and Zeng, Weili and Zhang, Richong and Zhu, Feng and Zhao, Rui},
  journal={arXiv preprint arXiv:2306.15195},
  year={2023}
}

@inproceedings{liu2024improved,
  title={Improved baselines with visual instruction tuning},
  author={Liu, Haotian and Li, Chunyuan and Li, Yuheng and Lee, Yong Jae},
  booktitle={Proceedings of the IEEE/CVF Conference on Computer Vision and Pattern Recognition},
  pages={26296--26306},
  year={2024}
}

@inproceedings{jiang2024hallucination,
  title={Hallucination augmented contrastive learning for multimodal large language model},
  author={Jiang, Chaoya and Xu, Haiyang and Dong, Mengfan and Chen, Jiaxing and Ye, Wei and Yan, Ming and Ye, Qinghao and Zhang, Ji and Huang, Fei and Zhang, Shikun},
  booktitle={Proceedings of the IEEE/CVF Conference on Computer Vision and Pattern Recognition},
  pages={27036--27046},
  year={2024}
}

@inproceedings{gunjal2024detecting,
  title={Detecting and preventing hallucinations in large vision language models},
  author={Gunjal, Anisha and Yin, Jihan and Bas, Erhan},
  booktitle={Proceedings of the AAAI Conference on Artificial Intelligence},
  volume={38(16)},
  pages={18135--18143},
  year={2024}
}

@article{liu2407paying,
  title={Paying more attention to image: A training-free method for alleviating hallucination in lvlms, 2024},
  author={Liu, Shi and Zheng, Kecheng and Chen, Wei},
  journal={URL https://arxiv. org/abs/2407.21771}, 
  year={2024}
}

@inproceedings{leng2024mitigating,
  title={Mitigating object hallucinations in large vision-language models through visual contrastive decoding},
  author={Leng, Sicong and Zhang, Hang and Chen, Guanzheng and Li, Xin and Lu, Shijian and Miao, Chunyan and Bing, Lidong},
  booktitle={Proceedings of the IEEE/CVF Conference on Computer Vision and Pattern Recognition},
  pages={13872--13882},
  year={2024}
}

@article{wang2024mllm,
  title={Mllm can see? dynamic correction decoding for hallucination mitigation},
  author={Wang, Chenxi and Chen, Xiang and Zhang, Ningyu and Tian, Bozhong and Xu, Haoming and Deng, Shumin and Chen, Huajun},
  journal={arXiv preprint arXiv:2410.11779},
  year={2024}
}

@inproceedings{liang2025mole,
  title={MoLE: Decoding by Mixture of Layer Experts Alleviates Hallucination in Large Vision-Language Models},
  author={Liang, Tian and Du, Yuetian and Huang, Jing and Kong, Ming and Chen, Luyuan and Li, Yadong and Chen, Siye and Zhu, Qiang},
  booktitle={Proceedings of the AAAI Conference on Artificial Intelligence},
  volume={39(18)},
  pages={18684--18692},
  year={2025}
}

@inproceedings{huang2024opera,
  title={Opera: Alleviating hallucination in multi-modal large language models via over-trust penalty and retrospection-allocation},
  author={Huang, Qidong and Dong, Xiaoyi and Zhang, Pan and Wang, Bin and He, Conghui and Wang, Jiaqi and Lin, Dahua and Zhang, Weiming and Yu, Nenghai},
  booktitle={Proceedings of the IEEE/CVF Conference on Computer Vision and Pattern Recognition},
  pages={13418--13427},
  year={2024}
}

@article{sarkar2025mitigating,
  title={Mitigating Hallucinations in Vision-Language Models through Image-Guided Head Suppression},
  author={Sarkar, Sreetama and Che, Yue and Gavin, Alex and Beerel, Peter A and Kundu, Souvik},
  journal={arXiv preprint arXiv:2505.16411},
  year={2025}
}

@inproceedings{jiang2025devils,
  title={Devils in middle layers of large vision-language models: Interpreting, detecting and mitigating object hallucinations via attention lens},
  author={Jiang, Zhangqi and Chen, Junkai and Zhu, Beier and Luo, Tingjin and Shen, Yankun and Yang, Xu},
  booktitle={Proceedings of the Computer Vision and Pattern Recognition Conference},
  pages={25004--25014},
  year={2025}
}

@inproceedings{yangunderstanding,
  title={Understanding and mitigating hallucination in large vision-language models via modular attribution and intervention},
  author={Yang, Tianyun and Li, Ziniu and Cao, Juan and Xu, Chang},
  booktitle={The Thirteenth International Conference on Learning Representations}, 
  year={2025}
}

@article{dai2023instructblip,
  title={Instructblip: Towards general-purpose vision-language models with instruction tuning},
  author={Dai, Wenliang and Li, Junnan and Li, Dongxu and Tiong, Anthony and Zhao, Junqi and Wang, Weisheng and Li, Boyang and Fung, Pascale N and Hoi, Steven},
  journal={Advances in neural information processing systems},
  volume={36},
  pages={49250--49267},
  year={2023}
}

@article{vaswani2017attention,
  title={Attention is all you need},
  author={Vaswani, Ashish and Shazeer, Noam and Parmar, Niki and Uszkoreit, Jakob and Jones, Llion and Gomez, Aidan N and Kaiser, {\L}ukasz and Polosukhin, Illia},
  journal={Advances in neural information processing systems},
  volume={30},
  year={2017}
}

@article{huang2025survey,
  title={A survey on hallucination in large language models: Principles, taxonomy, challenges, and open questions},
  author={Huang, Lei and Yu, Weijiang and Ma, Weitao and Zhong, Weihong and Feng, Zhangyin and Wang, Haotian and Chen, Qianglong and Peng, Weihua and Feng, Xiaocheng and Qin, Bing and others},
  journal={ACM Transactions on Information Systems},
  volume={43},
  number={2},
  pages={1--55},
  year={2025},
  publisher={ACM New York, NY}
}

@article{chen2024halc,
  title={Halc: Object hallucination reduction via adaptive focal-contrast decoding},
  author={Chen, Zhaorun and Zhao, Zhuokai and Luo, Hongyin and Yao, Huaxiu and Li, Bo and Zhou, Jiawei},
  journal={arXiv preprint arXiv:2403.00425},
  year={2024}
}

@article{sun2023aligning,
  title={Aligning large multimodal models with factually augmented rlhf},
  author={Sun, Zhiqing and Shen, Sheng and Cao, Shengcao and Liu, Haotian and Li, Chunyuan and Shen, Yikang and Gan, Chuang and Gui, Liang-Yan and Wang, Yu-Xiong and Yang, Yiming and others},
  journal={arXiv preprint arXiv:2309.14525},
  year={2023}
}

@article{wang2025tpc,
  title={TPC: Cross-Temporal Prediction Connection for Vision-Language Model Hallucination Reduction},
  author={Wang, Chao and Fu, Weiwei and Zhou, Yang},
  journal={arXiv preprint arXiv:2503.04457},
  year={2025}
}

@article{wang2025ascd,
  title={ASCD: Attention-Steerable Contrastive Decoding for Reducing Hallucination in MLLM},
  author={Wang, Yujun and Bi, Jinhe and Ma, Yunpu and Pirk, Soeren},
  journal={arXiv preprint arXiv:2506.14766},
  year={2025}
}

@article{jiang2025hicd,
  title={HICD: Hallucination-Inducing via Attention Dispersion for Contrastive Decoding to Mitigate Hallucinations in Large Language Models},
  author={Jiang, Xinyan and Ye, Hang and Zhu, Yongxin and Zheng, Xiaoying and Chen, Zikang and Gong, Jun},
  journal={arXiv preprint arXiv:2503.12908},
  year={2025}
}

@article{chen2025mitigating,
  title={Mitigating Hallucination of Large Vision-Language Models via Dynamic Logits Calibration},
  author={Chen, Jiahe and He, Jiaying and Shao, Qian and Chen, Qiyuan and Ying, Jiahe and Xu, Hongxia and Chen, Jintai and Zheng, Jianwei and Wu, Jian},
  journal={arXiv preprint arXiv:2506.21509},
  year={2025}
}

@article{tang2025mitigating,
  title={Mitigating Hallucinations via Inter-Layer Consistency Aggregation in Large Vision-Language Models},
  author={Tang, Kai and You, Jinhao and Ge, Xiuqi and Li, Hanze and Guo, Yichen and Huang, Xiande},
  journal={arXiv preprint arXiv:2505.12343},
  year={2025}
}

@inproceedings{lin2014microsoft,
  title={Microsoft coco: Common objects in context},
  author={Lin, Tsung-Yi and Maire, Michael and Belongie, Serge and Hays, James and Perona, Pietro and Ramanan, Deva and Doll{\'a}r, Piotr and Zitnick, C Lawrence},
  booktitle={European conference on computer vision},
  pages={740--755},
  year={2014},
  organization={Springer}
}

@article{rohrbach2018object,
  title={Object hallucination in image captioning},
  author={Rohrbach, Anna and Hendricks, Lisa Anne and Burns, Kaylee and Darrell, Trevor and Saenko, Kate},
  journal={arXiv preprint arXiv:1809.02156},
  year={2018}
}

@article{li2023evaluating,
  title={Evaluating object hallucination in large vision-language models},
  author={Li, Yifan and Du, Yifan and Zhou, Kun and Wang, Jinpeng and Zhao, Wayne Xin and Wen, Ji-Rong},
  journal={arXiv preprint arXiv:2305.10355},
  year={2023}
}

@article{gong2024damro,
  title={Damro: Dive into the attention mechanism of lvlm to reduce object hallucination},
  author={Gong, Xuan and Ming, Tianshi and Wang, Xinpeng and Wei, Zhihua},
  journal={arXiv preprint arXiv:2410.04514},
  year={2024}
}

@misc{logitlens,
    title = "interpreting GPT: the logit lens",
    author       = "{nostalgebraist}",
    howpublished = "\url{https://www.nasa.gov/nh/pluto-the-other-red-planet}",
    year         = 2020,
}

@inproceedings{he2023geometric,
  title={Geometric visual similarity learning in 3d medical image self-supervised pre-training},
  author={He, Yuting and Yang, Guanyu and Ge, Rongjun and Chen, Yang and Coatrieux, Jean-Louis and Wang, Boyu and Li, Shuo},
  booktitle={Proceedings of the IEEE/CVF Conference on Computer Vision and Pattern Recognition},
  pages={9538--9547},
  year={2023}
}

@article{chuang2023dola,
  title={Dola: Decoding by contrasting layers improves factuality in large language models},
  author={Chuang, Yung-Sung and Xie, Yujia and Luo, Hongyin and Kim, Yoon and Glass, James and He, Pengcheng},
  journal={arXiv preprint arXiv:2309.03883},
  year={2023}
}

@incollection{wilcoxon1992individual,
  title={Individual comparisons by ranking methods},
  author={Wilcoxon, Frank},
  booktitle={Breakthroughs in statistics: Methodology and distribution},
  pages={196--202},
  year={1992},
  publisher={Springer}
}

@article{li2023inference,
  title={Inference-time intervention: Eliciting truthful answers from a language model},
  author={Li, Kenneth and Patel, Oam and Vi{\'e}gas, Fernanda and Pfister, Hanspeter and Wattenberg, Martin},
  journal={Advances in Neural Information Processing Systems},
  volume={36},
  pages={41451--41530},
  year={2023}
}

@article{kang2025see,
  title={See what you are told: Visual attention sink in large multimodal models},
  author={Kang, Seil and Kim, Jinyeong and Kim, Junhyeok and Hwang, Seong Jae},
  journal={arXiv preprint arXiv:2503.03321},
  year={2025}
}

@inproceedings{kang2025your,
  title={Your large vision-language model only needs a few attention heads for visual grounding},
  author={Kang, Seil and Kim, Jinyeong and Kim, Junhyeok and Hwang, Seong Jae},
  booktitle={Proceedings of the Computer Vision and Pattern Recognition Conference},
  pages={9339--9350},
  year={2025}
}

@article{Qwen-VL,
  title={Qwen-VL: A Versatile Vision-Language Model for Understanding, Localization, Text Reading, and Beyond},
  author={Bai, Jinze and Bai, Shuai and Yang, Shusheng and Wang, Shijie and Tan, Sinan and Wang, Peng and Lin, Junyang and Zhou, Chang and Zhou, Jingren},
  journal={arXiv preprint arXiv:2308.12966},
  year={2023}
}

@article{uppaal2024model,
  title={Model editing as a robust and denoised variant of dpo: A case study on toxicity},
  author={Uppaal, Rheeya and Dey, Apratim and He, Yiting and Zhong, Yiqiao and Hu, Junjie},
  journal={arXiv preprint arXiv:2405.13967},
  year={2024}
}

@inproceedings{yang2025nullu,
  title={Nullu: Mitigating object hallucinations in large vision-language models via halluspace projection},
  author={Yang, Le and Zheng, Ziwei and Chen, Boxu and Zhao, Zhengyu and Lin, Chenhao and Shen, Chao},
  booktitle={Proceedings of the Computer Vision and Pattern Recognition Conference},
  pages={14635--14645},
  year={2025}
}
}


\clearpage
\newpage
\appendix
\section{Structure of The Appendix}
The appendix is structured as follows:

\cref{app:implement} details the datasets used in our validation experiments as well as the baseline methods for comparison. It also provides additional experiments that complement the main text, including throughput estimates across different models, ablation studies, and related analyses.

\cref{app:preliminary} details the statistical tools employed in this work, along with additional experimental instruments such as the Logit Lens.

\cref{app:framework} presents the overall framework of D-LEAF, demonstrating both the soundness and motivation of our proposed metrics, and further verifying their generalizability across model architectures. It also includes supplementary analyses on the role of layer priors in the algorithm and introduces the concept of visual processing layers.

\cref{app:case} provides qualitative case studies showcasing high-quality answers generated by models after applying our algorithm.

\cref{appcntdpo} provides the derivation of the DPO optimization objective and further elaborates on the conceptual connection between D-LEAF and DPO.

\section{Implementation Details}
\label{app:implement}
\subsection{Dataset} 
\label{app:dataset}
\paragraph{CHAIR.} The Caption Hallucination Assessment with Image Relevance (CHAIR) metric provides per‑image ground‑truth object annotations for image captioning, flagging any model‑generated object not in the reference set as a hallucination. 

\paragraph{POPE.} The Polling-based Object Probing Evaluation (POPE) evaluates hallucinations in visual question answering by querying “Is there a \textless{}object\textgreater{} in the image?” using three object‑sampling strategies:
    \begin{itemize}
        \item [\textbullet]Random: uniformly drawn from the full dataset.
        \item [\textbullet]Popular: selected from the most frequent objects.
        \item [\textbullet]Adversarial: chosen for strong semantic relevance to the image.
    \end{itemize}

\paragraph{MMHal-Benchmark.} MMHal‑Bench comprises 96 image–question pairs spanning 12 COCO‑derived object meta‑categories and eight question types (attributes, adversarial, comparison, counting, spatial relations, environment, holistic descriptions, and others), providing a rigorous testbed for evaluating model hallucination in challenging examples.

In addition, to evaluate the effectiveness of our method, we tested the number of tokens output per second by the model in each of the three models.

\subsection{Baselines}
\label{app:baseline}
\paragraph{Greedy Search and Nuclear Sampling.} Traditional decoding strategies that are widely used in sequence generation tasks.

\paragraph{SPIN and PAI.} SPIN \citep{sarkar2025mitigating} and PAI \citep{liu2407paying}: The latest SOTA approach leverages attention‑head mechanisms to effectively suppress hallucinations.

\paragraph{VCD.} VCD \citep{leng2024mitigating}: A technique that introduces noise into images to create amateur models for contrastive decoding.

\paragraph{DAMRO.} DAMRO \citep{gong2024damro}: This method leverages the ViT’s CLS token to selectively filter out high‑attention background outliers and eliminate their influence during decoding.

We used the parameters provided in the open source version of these methods.

\subsection{MMHal-Bench Evaluation}
\begin{figure}[t]
\centering
\includegraphics[width=1.0\linewidth]{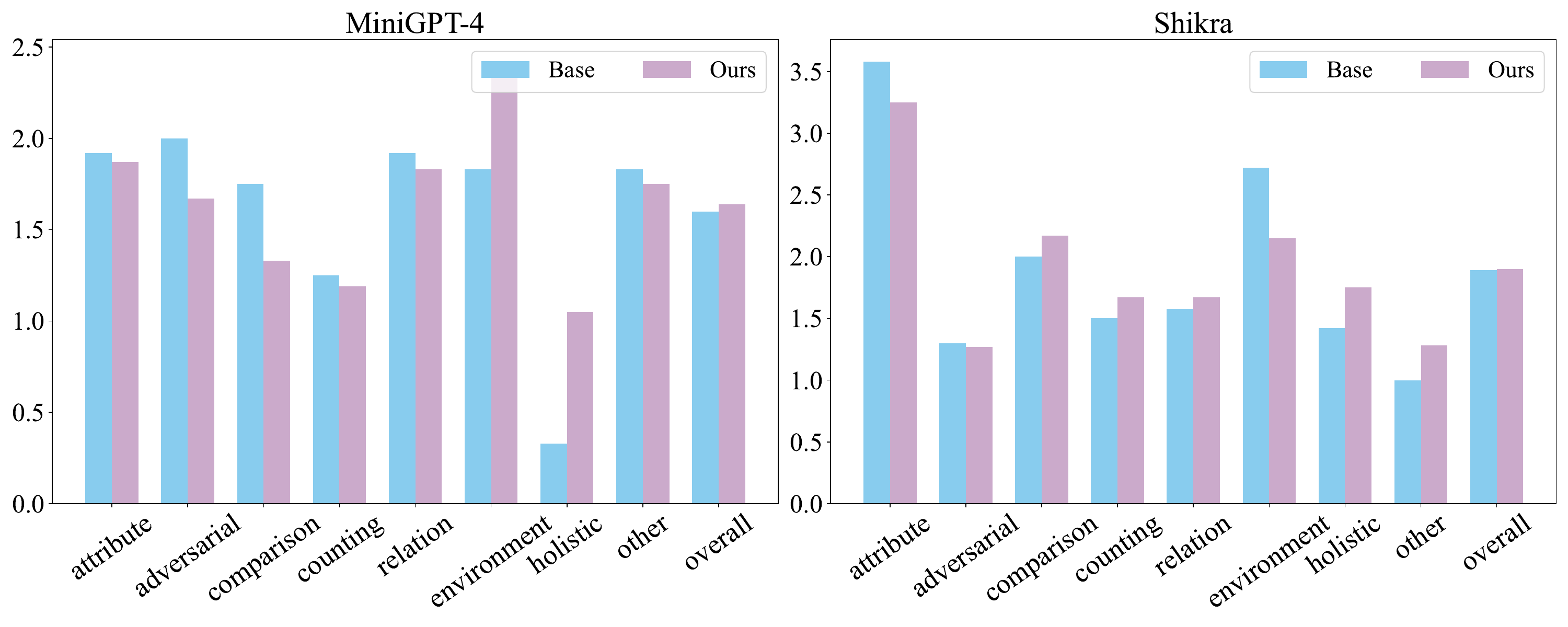}
\caption{MMHal-Bench Evaluation on MiniGPT-4 and Shikra.}
\label{mmbench_other}
\end{figure}

We present in \cref{mmbench_other} the performance of D-LEAF compared with the baseline on MiniGPT-4 and Shikra for the MMHal-Bench tasks. D-LEAF shows substantial improvement in descriptive tasks such as the holistic description task, and it also achieves a noticeable gain in the average performance across all task categories.
\label{appmmhal}

\subsection{Throughput Estimation}
\label{app:throughput}
\begin{figure}[!htbp]
\centering
\includegraphics[width=0.8\columnwidth]{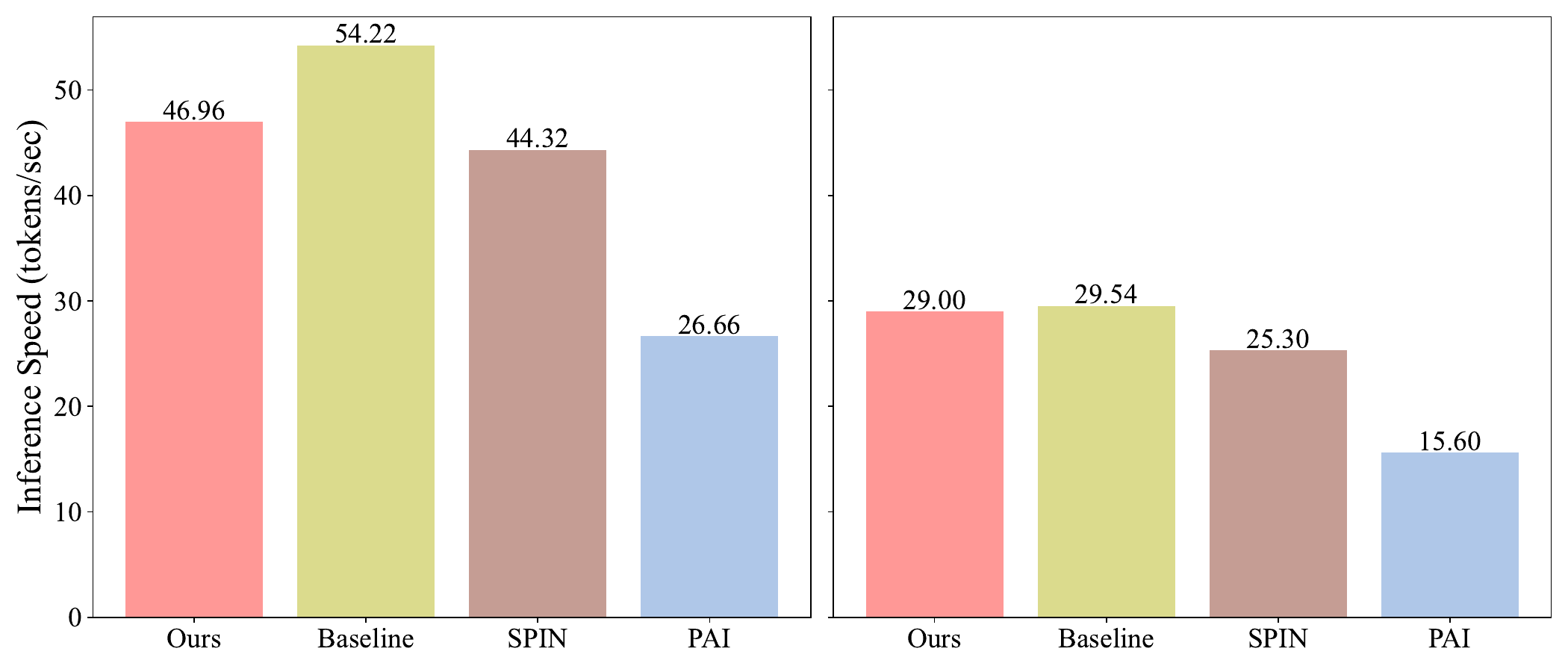}
\caption{Throughput comparison with existing methods given by the number of tokens generated per second in MiniGPT-4 (left) and Shikra (right).}
\label{minigpt4_shikra_through}
\end{figure}

In the main text, we demonstrated that on LLaVA our algorithm achieves throughput closest to greedy search among all methods. To further validate its effectiveness, we measured throughput on MiniGPT-4 and Shikra. As shown in \cref{minigpt4_shikra_through}, our approach still incurs the smallest throughput degradation while maintaining high hallucination suppression rates and preserving output detail.

\subsection{Ablation Study}
\label{app:abl}
D-LEAF is a two-stage localization–correction algorithm for hallucination suppression. In the first stage, localization is performed by comparing the Layer Image Attention Entropy (LIAE) with the current Best Attention Score (BAS) to identify abnormal layers. This process does not require any additional hyperparameters, though layer priors, $L$, can be applied during this phase, which restricts localization to specific layers and improve detection accuracy. In the second stage, correction is carried out by ranking the attention heads within the identified abnormal layers using the Image Attention Focus (IAF). The lowest-performing $n$ heads are then selected and refined through a mixing adjustment with a correction coefficient $\gamma$.

\cref{lay_abl} shows that incorporating Layer Priors, $L$, into our algorithm yields a 10$\%$ improvement in hallucination reduction compared to the variant without layer priors. Regardless of whether layer priors are applied, our method consistently achieves the best performance among hallucination suppression algorithms. However, we observe that while hallucinations decrease, the model’s F1 score also drops—an issue similarly reported in other attention-head-based suppression methods \citep{liu2407paying, sarkar2025mitigating}. In \cref{app:visual}, we further evaluate this phenomenon and find that introducing layer priors can mitigate this decline, enabling hallucination reduction while maintaining or even improving F1. Nonetheless, a fundamental trade-off remains between the two.

\begin{table*}[!htbp]
  \centering
  \caption{Ablation Study of Layer Prior. The best result is highlighted in bold, and the second-best is underlined.}
  \label{lay_abl}
  \resizebox{0.7\textwidth}{!}{%
  \begin{tabular}{ccccc}
   \toprule
    Model        & $L$  & C$_S$ & C$_I$ & F1\\
    \midrule
    \multirow{3}{*}{LLAVA} 
                 & baseline  & 47.08 $\pm 1.54$ & 13.00 $\pm 0.59$ & \textbf{77.22 $\pm 0.66$}\\
                 & with L    & \textbf{23.44} $\pm 2.63$ & \textbf{6.72 $\pm 0.49$}  & \underline{74.88 $\pm 1.51$}\\
                 & without L & \underline{26.20 $\pm 2.63$} & \underline{8.30 $\pm 1.32$} & 74.28 $\pm 0.81$\\
    \midrule
    \multirow{3}{*}{MiniGPT-4}
                 & baseline  & 34.00 $\pm 1.98$ & 10.82 $\pm 0.59$ & \textbf{69.60 $\pm 0.99$}\\
                 & with L    & \textbf{11.56 $\pm 1.69$}    & \textbf{4.72 $\pm 0.95$}     & 66.06 $\pm 0.99$\\
                 & without L & \underline{14.60 $\pm 1.51$} & \underline{6.76  $\pm 1.13$} & \underline{66.52  $\pm 0.61$}\\
    \midrule
    \multirow{3}{*}{Instrutblip}
                 & baseline  & 48.12 $\pm 2.98$             & 14.18 $\pm 1.12$            & \textbf{73.92 $\pm 0.82$}\\
                 & with L    & \underline{22.28 $\pm 1.32$} & \textbf{7.82 $\pm 2.88$}    & 70.06 $\pm 0.80$\\
                 & without L & \textbf{22.44 $\pm 2.75$}    & \underline{8.48 $\pm 5.93$} & \underline{70.74 $\pm 1.13$}\\
    \bottomrule
  \end{tabular}
}
\end{table*}

\begin{figure}[t]
\centering
\includegraphics[width=1.0\columnwidth]{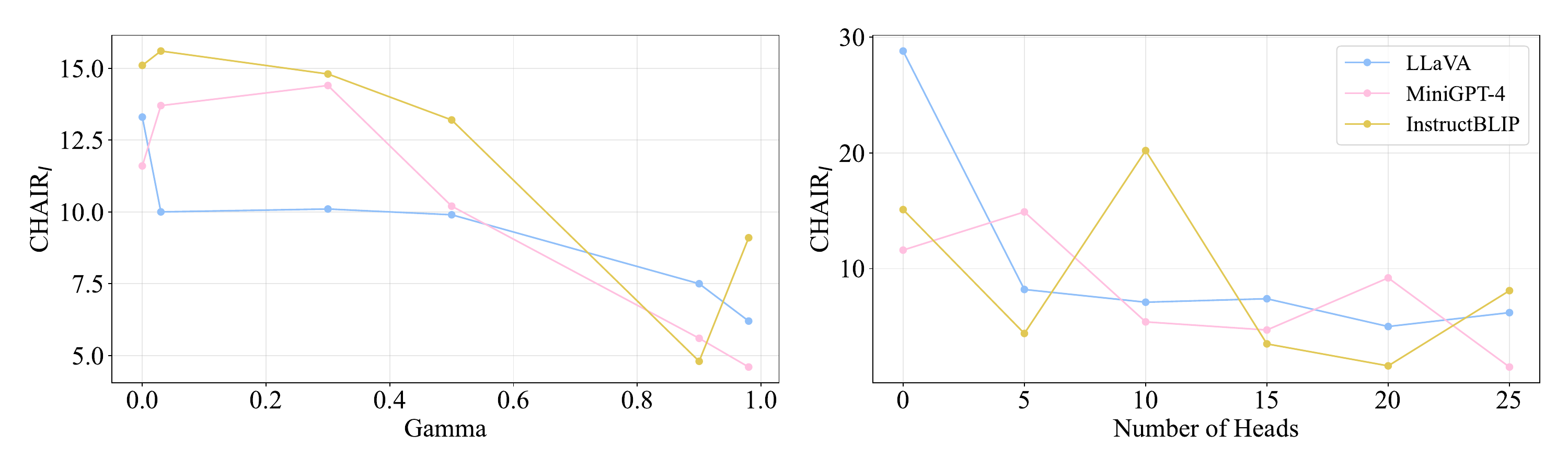}
\vspace{-13pt}
\caption{Ablation Study results with CHAIR$_I$ for hyperparamter $\gamma$ and $n$.}
\label{CI_abl}
\vspace{-13pt}
\end{figure}

\begin{figure}[t]
\centering
\includegraphics[width=1.0\columnwidth]{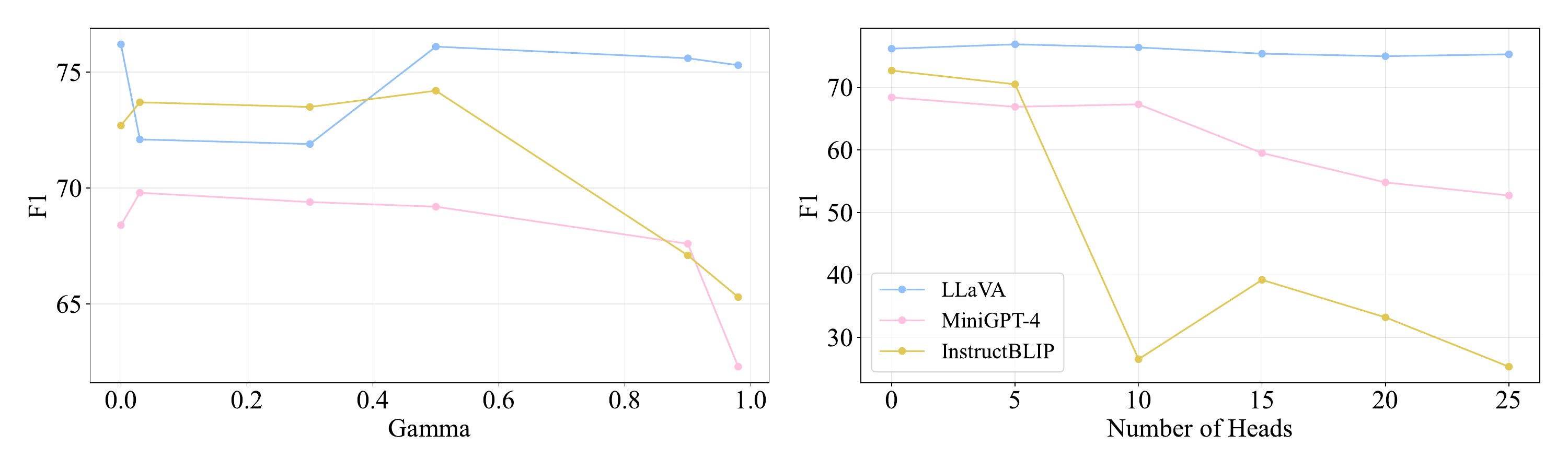}
\vspace{-13pt}
\caption{Ablation Study results with F1 for hyperparamter $\gamma$ and $n$.}
\label{F1_abl}
\vspace{-13pt}
\end{figure}

In addition to the analysis in the main text on the impact of the mixing coefficient $\gamma$ and the number of suppressed attention heads $n$ on $\mathrm{CHAIR}_S$, we further report here the influence of these hyperparameters on $\mathrm{CHAIR}_I$ and F1, as shown in \cref{CI_abl} and \cref{F1_abl}. The results demonstrate the strong robustness of our method: across a wide range of $\gamma$ values (0.03 to 0.98), our algorithm consistently outperforms the baseline. We further observe that increasing $\gamma$ leads to a significant improvement in hallucination suppression across all three models. However, overly large $\gamma$ values result in a drop in F1 score, whereas appropriately chosen $\gamma$ achieves a favorable balance—reducing hallucinations while maintaining high F1 performance.

As for the impact of the number of corrected attention heads, $n$, on hallucination reduction, we find that in models such as InstructBLIP, which leverage a learnable querying transformer to establish vision–language connections with only 32 image tokens as MLLM input, correcting even a small subset of attention heads achieves strong suppression performance. However, as the number of corrected heads increases, the model’s output capability deteriorates significantly: the F1 score drops sharply and the generated responses become markedly shorter. We attribute this to excessive correction disrupting the model’s normal output dynamics, causing it to prefer shorter responses as a way of avoiding hallucinated tokens. Interestingly, contrary to this trend, in LLaVA, which employs an MLP to map the vision branch output into 576 image embeddings, the number of corrected attention heads does not substantially affect model performance.

\label{app:abl}

\section{Preliminary}
\label{app:preliminary}
In this section, we provide a detailed overview of the analysis tools employed, such as LogitLens and the Wilcoxon signed-rank test, and the forward process of MLLMs.

\subsection{Model Forward Process}
The language decoder module here consists of multiple transformer components \citep{vaswani2017attention}. Each transformer block comprises two sublayers: a multi-head attention (MHA) mechanism and a feed-forward network (FFN). MHA begins by taking the combined text and image embedding vectors \text{$X \in R^{N \times d}$} as input, projecting them into the query \text{$(Q)$}, key \text{$(K)$}, and value \text{$(V)$} spaces, and then computing the output of the MHA module. The output of MHA is fed into the FFN module, and the final output of the current encoder block is obtained through residual flow, as shown in \cref{eq1} and \cref{eq2}.

\begin{equation}
\label{eq1}
    x_n^{mid,l} = \sum_{h=1}^{H}\text{Attn}^{(l,h)}(X_{\leq n}^{l-1})W_U
\end{equation}
\begin{equation}
\label{eq2}
    f(x) = \sum_{l=1}^{L}x_n^{mid,l} + \sum_{l=1}^{L}\text{FFN}^l(x_n^{mid,l})W_U+x_nW_U
\end{equation}

Therefore, we are able to suppress hallucinations by making corrections in the MHA module during the forward process without modifying the model architecture or adding additional training. 

\subsection{LogitLens}
\label{logitlens}
LogitLens \citep{logitlens} is an interpretability technique that directly maps each hidden state $x^l$ to the model’s vocabulary distribution by first applying the LayerNorm transformation and then projecting through the unembedding matrix $W_U$, as shown in \cref{logit}.
\begin{equation}
\label{logit}
    \text{LogitLens}(x^l) = \text{LayerNorm}(x^l)W_U
\end{equation}

We used LogitLens to analyze the probability curves of the model’s target logit under both selective and indiscriminate correction, which enabled us to recognize that indiscriminate correction did not genuinely take effect in certain scenarios.

\subsection{Wilcoxon Signed-rank Test}
\label{Wilcoxon}
The Wilcoxon signed-rank test is a non-parametric method for assessing whether the median difference between paired samples is zero. Given paired observations \((x_i, y_i)\), we compute the differences \[d_i = x_i - y_i,\] exclude any zero differences, and rank the remaining absolute values \(\lvert d_i\rvert\) to obtain ranks \(R_i\). The test statistic is then defined as
\begin{equation}\label{eq15}
W = \sum_{i=1}^{n} \operatorname{sign}(d_i)\,R_i.
\end{equation}
Under the null hypothesis that the distributions of \(x_i\) and \(y_i\) are identical, \(W\) has a known sampling distribution, from which we derive a two-sided \(p\)-value to determine significance.

\subsection{Isotonic Regression}
\label{isotonic}
Isotonic regression is a non-parametric technique for fitting a monotonic (non-decreasing) function to a set of paired observations \((x_i, y_i)\). It estimates values \(f_i\) by solving
\begin{equation}\label{eq16}
\min_{f_1 \le f_2 \le \cdots \le f_n} \sum_{i=1}^n w_i\,(y_i - f_i)^2,
\end{equation}
subject to the ordering constraints \(f_i \le f_{i+1}\), where \(w_i\) are optional non-negative weights. This problem is efficiently solved using the Pool Adjacent Violators Algorithm (PAVA), which produces a piecewise-constant fit that enforces the desired monotonic relationship.

\subsection{Spearman Correlation Coefficient}
\label{spearman}
The Spearman correlation coefficient $\rho$ is a non-parametric measure of rank correlation that evaluates the strength and direction of a monotonic relationship between two variables. Given paired observations $(x_i, y_i)$ for $i=1,\dots,n$, we first convert them to ranks $R(x_i)$ and $R(y_i)$, and then compute
\begin{equation}\label{eq17}
\rho = \frac{\sum_{i=1}^n \bigl(R(x_i) - \overline{R_x}\bigr)\,\bigl(R(y_i) - \overline{R_y}\bigr)}%
{\sqrt{\sum_{i=1}^n\bigl(R(x_i) - \overline{R_x}\bigr)^2}\;\sqrt{\sum_{i=1}^n\bigl(R(y_i) - \overline{R_y}\bigr)^2}}\,,
\end{equation}
where $\overline{R_x} = \frac{1}{n}\sum_{i=1}^n R(x_i)$ and $\overline{R_y} = \frac{1}{n}\sum_{i=1}^n R(y_i)$.  
Alternatively, when there are no tied ranks, it can be expressed as
\[
\rho = 1 - \frac{6\sum_{i=1}^n d_i^2}{n(n^2-1)}, 
\quad d_i = R(x_i) - R(y_i).
\]
The coefficient ranges from $-1$ (perfect negative correlation) to $+1$ (perfect positive correlation), with $\rho=0$ indicating no monotonic association.

\section{Details of D-LEAF Framework}
\label{app:framework}
In this section, we first provide an an empirical analysis on the metrics in D-LEAF, followed by verification of the generalisability of various metrics in D-LEAF under other model architectures (Shikra) and the overall process of the D-LEAF algorithm.

\subsection{Empirical Analysis on Entropy and Focus}
Previous studies have suggested that insufficient and overly dispersed visual-stream attention is one of the primary causes of hallucination in MLLMs. To validate this claim, we examine Shikra and MiniGPT-4, comparing the degree of attention to image regions and the entropy of attention distributions across different layers when the models generate hallucinated versus factual tokens.

\begin{figure}[!htbp]
\centering
\includegraphics[width=1.0\columnwidth]{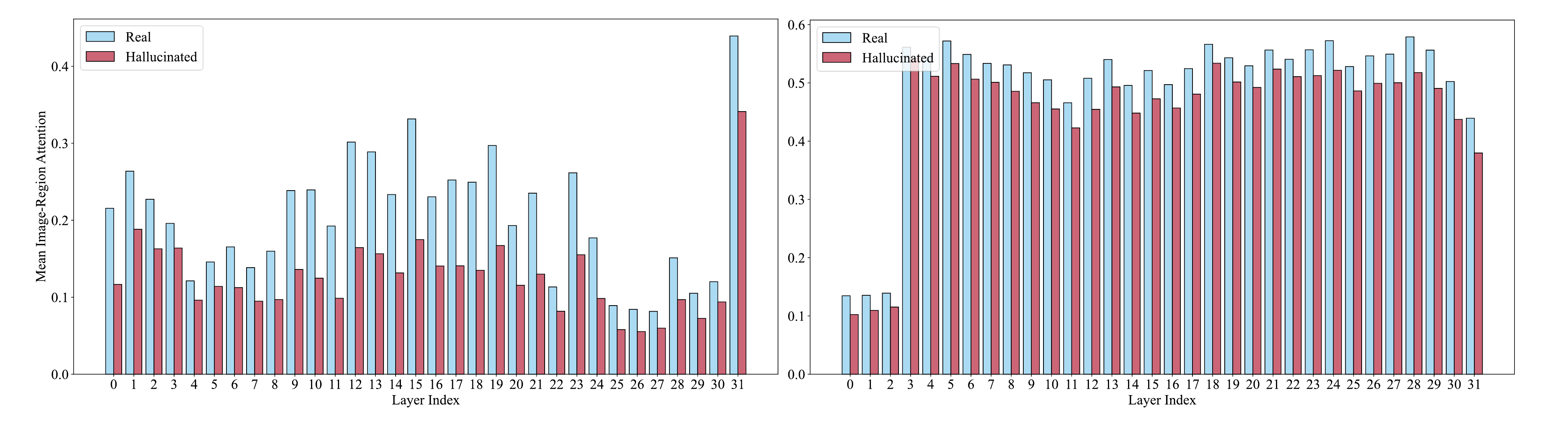}
\caption{Comparison of attention scores for real words and hallucinated words in the image region in MiniGPT-4 (left) and Shikra (right).}
\label{attn_cmp}
\end{figure}

We randomly sampled 500 images from the COCO2014 validation set and, for each image, extracted the model’s attention matrices when generating hallucinated versus ground‑truth tokens, respectively. We then computed the attention scores over the image region for both cases. The results are shown in \cref{attn_cmp}. We observe that, in both models, across all layers, image‑region attention for ground‑truth tokens is higher than hallucinated tokens.

\begin{figure}[!htbp]
\centering
\includegraphics[width=1.0\columnwidth]{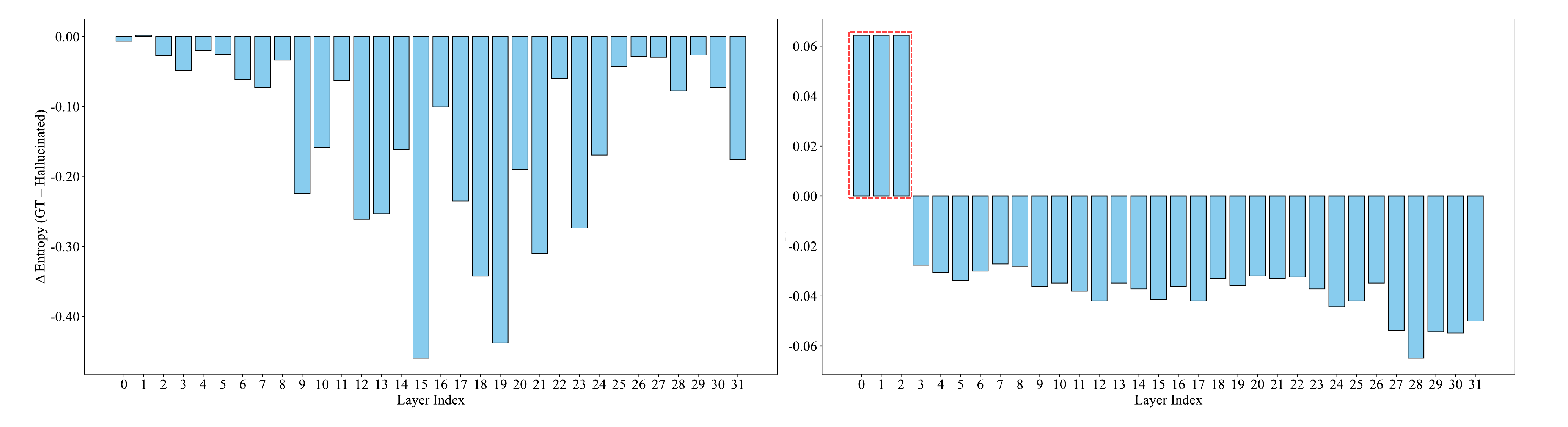}
\caption{Layer‐Wise Entropy Difference: Ground‑Truth minus Hallucinated in MiniGPT-4 (left) and Shikra (right).}
\label{entropy_cmp}
\end{figure}

Similarly, after applying softmax normalization to the attention matrices for ground‑truth and hallucinated tokens, we computed the mean image‑region entropy across all attention heads at each layer and plotted their per‑layer difference as a bar chart in \cref{entropy_cmp}. Because the raw entropy values are very close, we scaled the y‑axis by a factor of 10\textsuperscript{3}. We could observe that, across all layers, image‑region entropy for real tokens is lower than hallucinated tokens. However, in some architectures, like shikra, we observe that in certain layers the average entropy is actually higher for real words than for hallucinated terms. This further confirms that the indiscriminate modification of attention heads, as previously discussed, is suboptimal.

\subsection{Metric Validation and Comparison}
\label{app:metric_val}
In D-LEAF, we employ the Layer Image Attention Entropy (LIAE) to detect abnormal layers. Once abnormal layers are identified, we rank attention heads using the Image Attention Focus (IAF) to select those requiring correction. In this section, we present four sets of experiments demonstrating that for abnormal layer detection, using LIAE alone outperforms either IAF or a combined metric, whereas for head localization, using IAF alone yields better performance than either IAE or the combined approach.

Inspired by the previous section, we propose layer image attention as a comparison metric for anomaly layer detection.
\begin{equation}
\label{eqliaf}
    \text{LIAF}^{(l)} = \sum_{n=1}^{N} \text{MAM}_n^{(l)}
\end{equation}

\begin{figure}[!htbp]
\centering
\includegraphics[width=1.0\columnwidth]{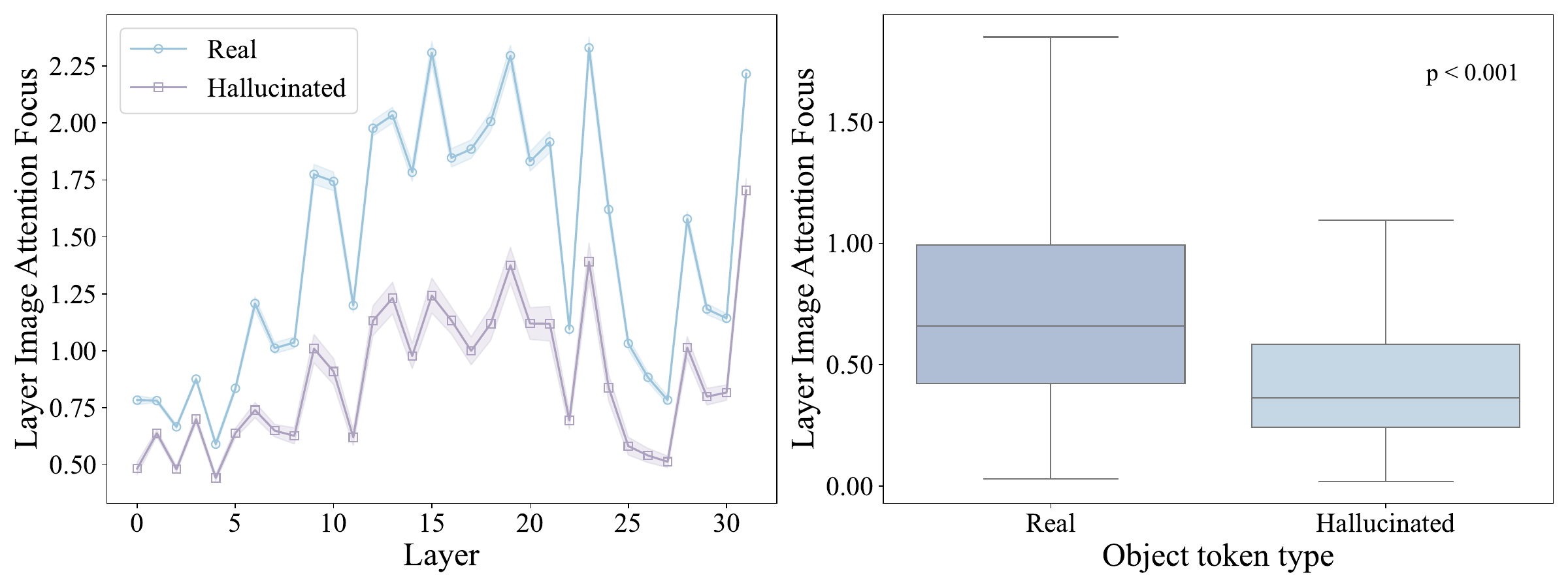}
\caption{LIAF change curves (left) and distributions , together with the Wilcoxon signed‑rank test results (right) for hallucinated versus real words generated by MiniGPT-4.}
\label{minigpt4_mech}
\end{figure}

To evaluate the significance of the metric, which independent and non‑normally distributed across real and hallucinated tokens, we apply the Wilcoxon signed‑rank test \citep{wilcoxon1992individual} as in main text. With $p < 0.001$, in \cref{minigpt4_mech} we confidently observe in MiniGPT-4 that hallucinated tokens exhibit significantly lower LIAF compared to real tokens.

\begin{figure}[!htbp]
\centering
\includegraphics[width=1.0\columnwidth]{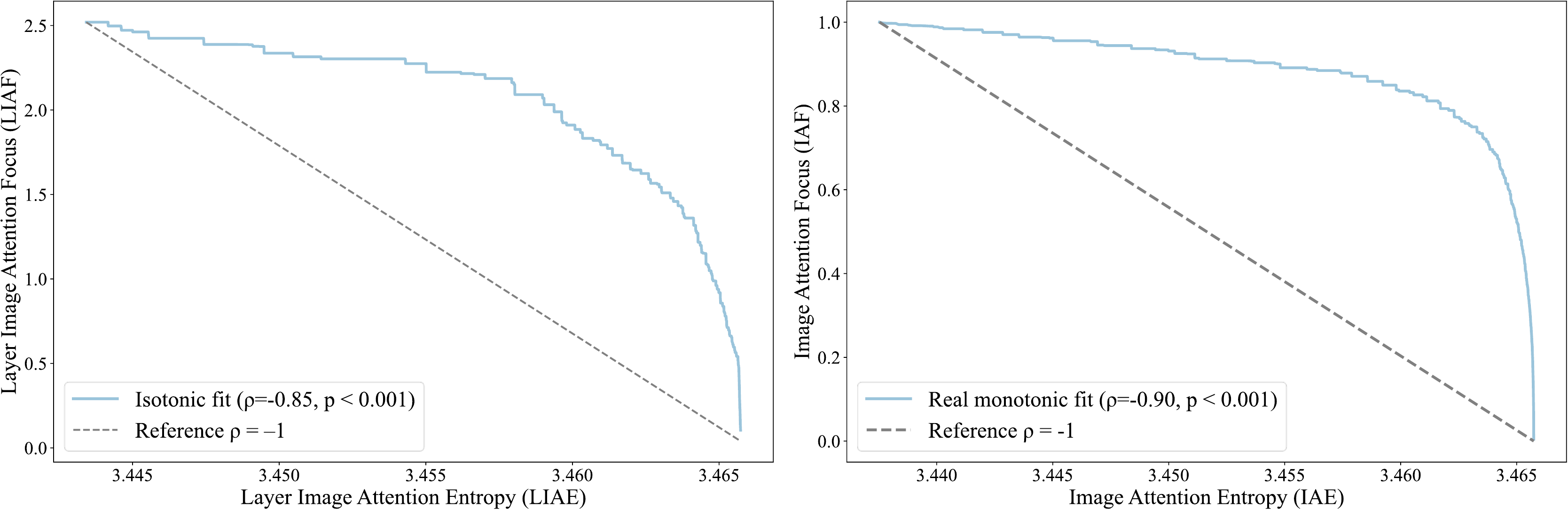}
\vspace{-10pt}
\caption{Isotonic Regression fit of LIAE against LIAF (left) and IAE againest IAF (right) in MiniGPT-4.}
\label{iso_minigpt4}
\vspace{-12pt}
\end{figure}

While we have already confirmed the individual effectiveness of LIAE and LIAF, combining them necessitates conducting a correlation analysis between the two metrics. Since these two metrics do not conform to a normal distribution, we computed the Spearman correlation coefficient between LIAE and LIAF, obtaining $\rho = –0.85$ with $p < 0.001$, which indicates that there is a high negative correlation between LIAE and LIAF. To further characterize their relationship, we fitted an isotonic regression—shown as the solid blue curve in \cref{iso_minigpt4} left closely follows, yet remains slightly above, the idealized gray dashed line denoting perfect negative correlation.  We obtained the same fitting results for attention head localization metrics: Image Attention Entropy (IAE) and Image Attention Focus (IAF) as illustrated in \cref{iso_minigpt4} right.

Based on the above experiments, we can conclude that there is a strong negative correlation between LIAE and LIAF. Therefore, we propose Layer Image Attention Score (LIAS) as a comprehensive indicator for hallucination detection: 
\begin{equation}
\label{eq6}
    \text{LIAS}^{(l)} = \alpha\ \text{LIAE}^{(l)} - (1-\alpha)\ \text{LIAF}^{(l)}
\end{equation}

To comprehensively evaluate the capability of the three proposed metrics in detecting abnormal layers, we conducted experiments on the CHAIR dataset using LLaVA, MiniGPT-4, and Shikra, as shown in the \cref{abl_alpha}. The results demonstrate that selecting LIAE alone as the primary indicator yields the strongest hallucination suppression. However, as LIAF is incorporated, the suppression effect gradually diminishes: once $\alpha$ exceeds 0.5, all three models produce identical detection results. We attribute this to the substantial numerical disparity between LIAE and LIAF, which causes LIAF to increasingly dominate in the mixed metric, whereas LIAE is inherently more sensitive to abnormalities. Therefore, we use LIAE exclusively during detection.
\begin{table}[!htbp]
  \centering
  \caption{Ablation study of detection coefficient $\alpha$. The best result is highlighted in bold, and the second-best is underlined.}
  \label{abl_alpha}
  \resizebox{0.45\textwidth}{!}{
  \begin{tabular}{c ccc ccc ccc}
    \toprule
    \multirow{2}{*}{$\alpha$} 
      & \multicolumn{3}{c}{LLaVA} 
      & \multicolumn{3}{c}{MiniGPT-4} 
      & \multicolumn{3}{c}{Shikra} \\
    \cmidrule{2-10}
      & C$_S$ & C$_I$ & F1
      & C$_S$ & C$_I$ & F1
      & C$_S$ & C$_I$ & F1 \\
   \midrule
    0.0 & \textbf{20.6} & \textbf{6.2}  & 75.3
        & \textbf{12.6} & \textbf{5.4}  & \underline{67.3}
        & \textbf{25.2} & \textbf{10.2} & 62.5 \\
    0.3 & 33.0 & 12.7 & \textbf{76.9}
        & \underline{35.4} & \underline{10.7} & 69.0
        & 35.2 & \underline{12.7} & \textbf{67.3} \\
    0.5 & \underline{32.0} & \underline{10.3} & \underline{75.0}
        & \underline{35.4} & \underline{10.7} & \textbf{69.5}
        & 35.4 & 13.4 & \underline{66.9} \\
    0.7 & \underline{32.0} & \underline{10.3} & \underline{75.0}
        & \underline{35.4} & \underline{10.7} & \textbf{69.5}
        & \underline{35.0} & 13.2 & \textbf{67.3} \\
    1.0 & \underline{32.0} & \underline{10.3} & \underline{75.0}
        & \underline{35.4} & \underline{10.7} & \textbf{69.5}
        & \underline{35.0} & 13.2 & \textbf{67.3} \\
   \bottomrule
  \end{tabular}
  }
\end{table}

In the correction stage, we additionally introduce two comparative metrics, namely the Image Attention Entropy (IAE) and the Image Attention Score (IAS). 
\begin{equation}\label{iae}
\text{IAE}_h^{(l)} = - \sum_{n=1}^{N} \text{P}(A_{h, n}^{(l)})\text{logP}(A_{h, n}^{(l)})
\end{equation}
\begin{equation}\label{ias}
\text{IAS}_h^{(l)} = \beta\ \text{IAF}_h^{(l)} + (1-\beta)\ \text{IAE}_h^{(l)}
\end{equation}

\begin{figure}[!htbp]
\centering
\includegraphics[width=1.0\columnwidth]{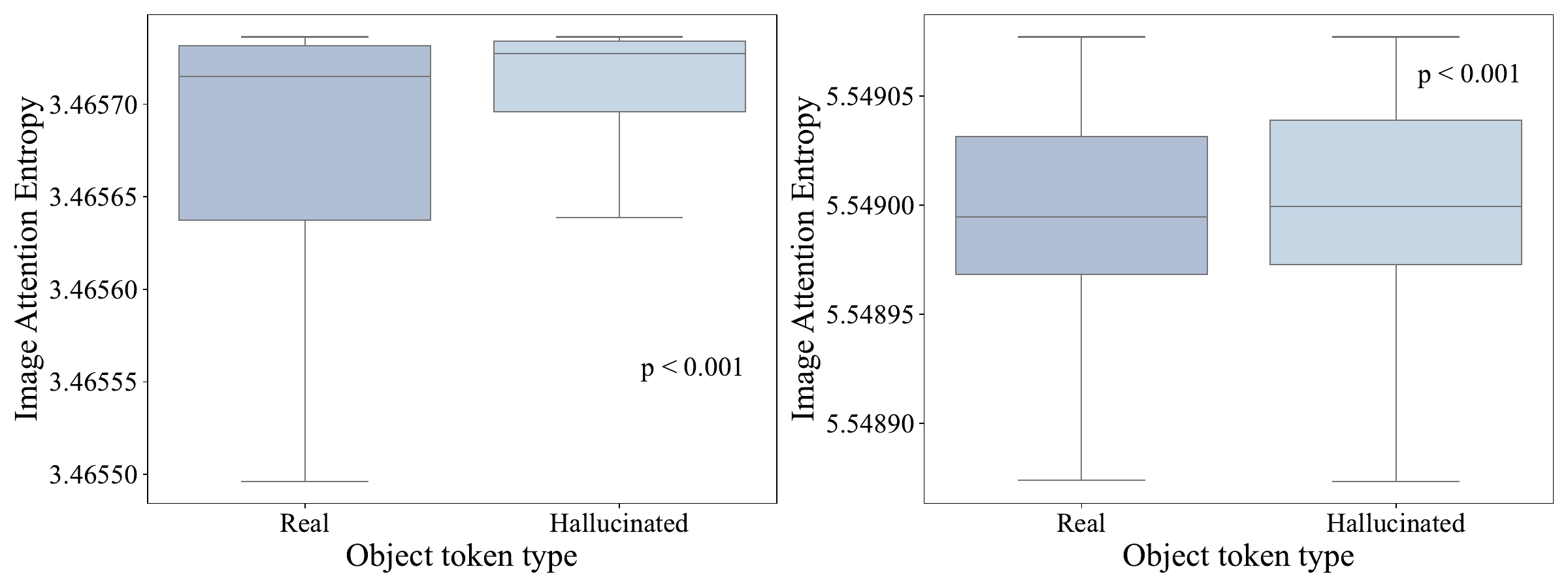}
\vspace{-10pt}
\caption{IAE distributions , together with the Wilcoxon signed‑rank test results for hallucinated versus real words generated by MiniGPT-4 (left) and Shikra (right).}
\label{minigpt4_shikra_iae}
\vspace{-12pt}
\end{figure}

We visualize the distributional differences of IAE when the model generates hallucinated versus factual tokens, as shown in \cref{minigpt4_shikra_iae}. The results reveal that the discrepancies across attention heads are extremely subtle, appearing only beyond the fifth decimal place. We further repeat the CHAIR experiment and find that once IAE is incorporated, the correction process leads the model to malfunction, as it mistakenly identifies and modifies the wrong attention heads. For these reasons, we use IAF exclusively during correction.

\subsection{Evaluation of Metric Generalizability}
\label{app:genral}
\begin{figure}[!htbp]
\centering
\includegraphics[width=1.0\columnwidth]{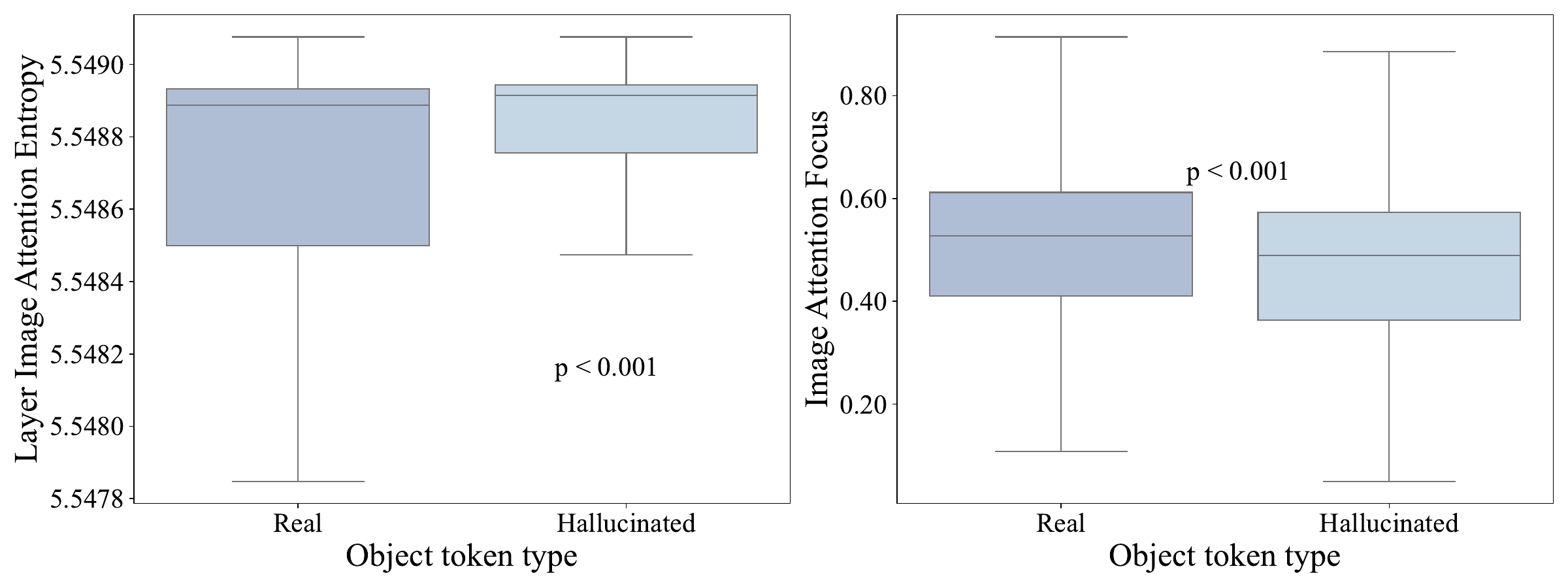}
\caption{The distributions of LIAE (left) and IAF (right) for hallucinated versus real words, together with the Wilcoxon signed‑rank test results in Shikra.}
\label{shikra_liae_iaf}
\end{figure}

To further verify the validity and effectiveness of our proposed metrics across different model architectures, we conducted supplementary experiments on other model architecture and analyzed the results. We first plot the distributions of LIAE and IAF in Shikra like the main context in \cref{shikra_liae_iaf}.

\begin{figure}[!htbp]
\centering
\includegraphics[width=1.0\columnwidth]{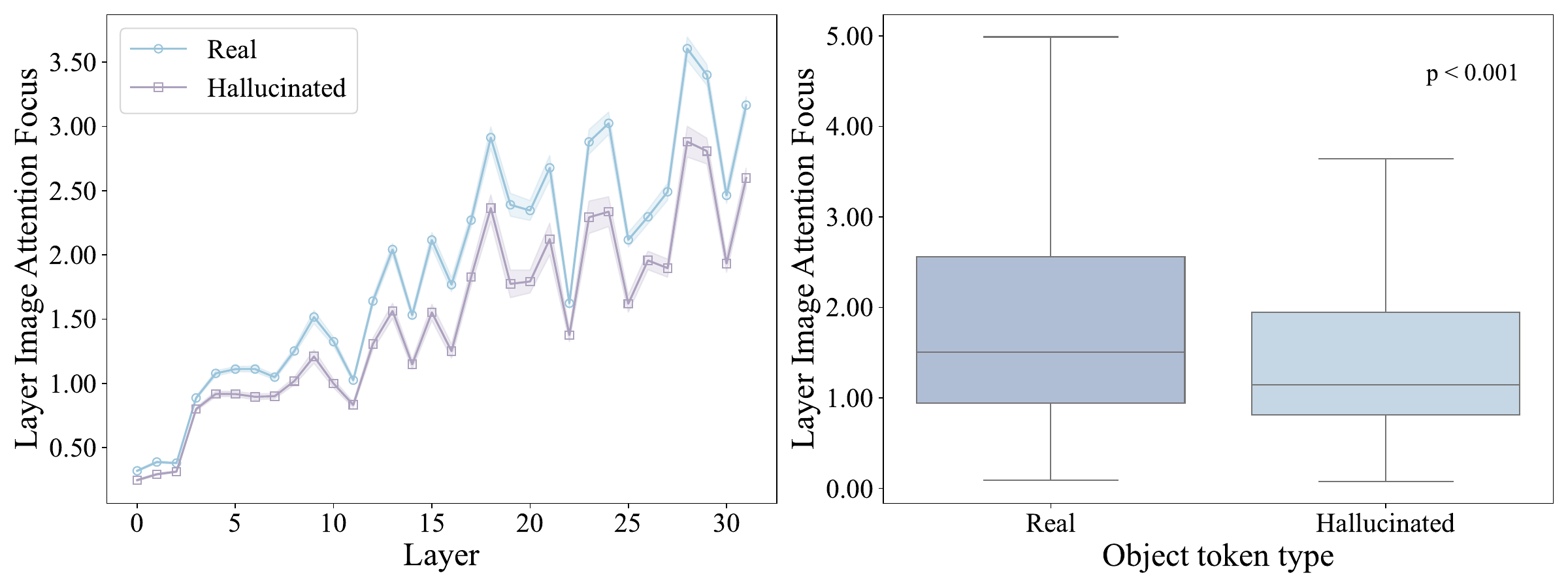}
\caption{LIAF change curves (left) and distributions , together with the Wilcoxon signed‑rank test results (right) for hallucinated versus real words generated by Shikra.}
\label{shikra_liaf}
\end{figure}

Additionally, we plotted the LIAF curve and distributions in \cref{shikra_liaf} and performed Wilcoxon signed‑rank tests on both metrics for hallucinated versus ground‑truth tokens. At p $<$ 0.001, hallucinated tokens exhibit significantly lower LIAF than ground‑truth tokens.

\begin{figure}[!htbp]
\centering
\includegraphics[width=1.0\columnwidth]{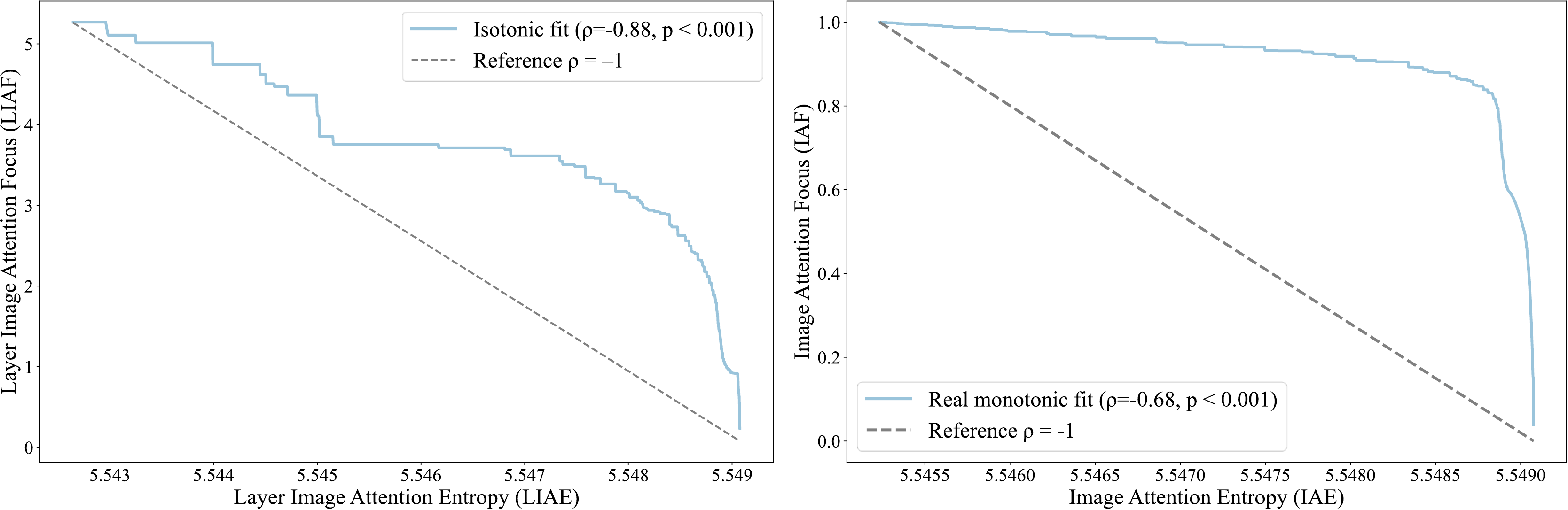}
\caption{Isotonic Regression fit of LIAE against LIAF (left) and IAE againest IAF (right) in Shikra.}
\label{iso_shikra}
\end{figure}

Furthermore, we evaluated the correlation between LIAE and LIAF and fitted it with isotonic regression, as shown in \cref{iso_shikra} left. The gray dashed line denotes a perfect negative correlation, and the blue curve represents the fitted regression; we observe a correlation of –0.88 at p$<$0.001.

After confirming that LIAE and LIAF generalize well for identifying anomalous layers, we further evaluated the generalizability of Image Attention Entropy (IAE) and Image Attention Focus (IAF) for pinpointing anomalous attention heads. The distribution and significance test of IAE is presented in \cref{minigpt4_shikra_iae}: for each attention head, Image Attention Entropy (IAE) in ground‑truth tokens is significantly lower than in hallucinated tokens (p $<$ 0.001).

Similarly, we applied isotonic regression fitting; as shown in \cref{iso_shikra} right, although our curve does not perfectly align with the gray dashed line denoting a perfect negative correlation, it still exhibits a clear negative trend, with a correlation coefficient of –0.68 (p$<$0.001).

In summary, on Shikra we observe the consistent patterns as in MiniGPT-4: ground‑truth tokens exhibit significantly lower LIAE and IAE and significantly higher LIAF and IAF compared to hallucinated tokens. Furthermore, both the LIAE–LIAF and IAE–IAF pairs show strong negative correlations, supporting their joint use as detection metrics for anomalous layers and heads.

\subsection{Visual Processing Layer}
\label{app:visual}
In the \cref{app:abl}, we conducted an ablation study on the Layer Prior and found that even without incorporating it, our algorithm already achieves state-of-the-art performance. However, introducing the layer prior further reduces hallucination rates. In this section, we provide a more fine-grained analysis of the Layer Prior and introduce the concept of visual processing layers.

We first evaluated the effect of varying the number of corrected layers on three models: LLaVA, MiniGPT-4, and Shikra, measuring changes in hallucination rate and F1 score, as shown in \cref{tab4}. The results demonstrate that regardless of the partitioning strategy, our method consistently achieves a substantial reduction in hallucination rates. Moreover, by tuning the number of corrected layers, our approach is able to maintain high F1 scores while suppressing hallucinations, further confirming both the robustness and strong transferability of our algorithm.

In addition, our experiments reveal that applying corrections within layers 0–25 often yields the most substantial reduction in hallucinations, albeit at the cost of some F1 degradation. We hypothesize that this effect arises because the algorithm is restricted to the visual processing stages: while the model becomes highly effective at distinguishing objects during visual processing, in the subsequent language generation phase it tends to produce shorter outputs to avoid hallucinated tokens, thereby leading to a decline in F1 score. To confirm this, we ran a simple experiment to confirm that all three architectures integrate visual features primarily in layers 0–25.

\begin{figure}[!htbp]
\centering
\includegraphics[width=0.8\columnwidth]{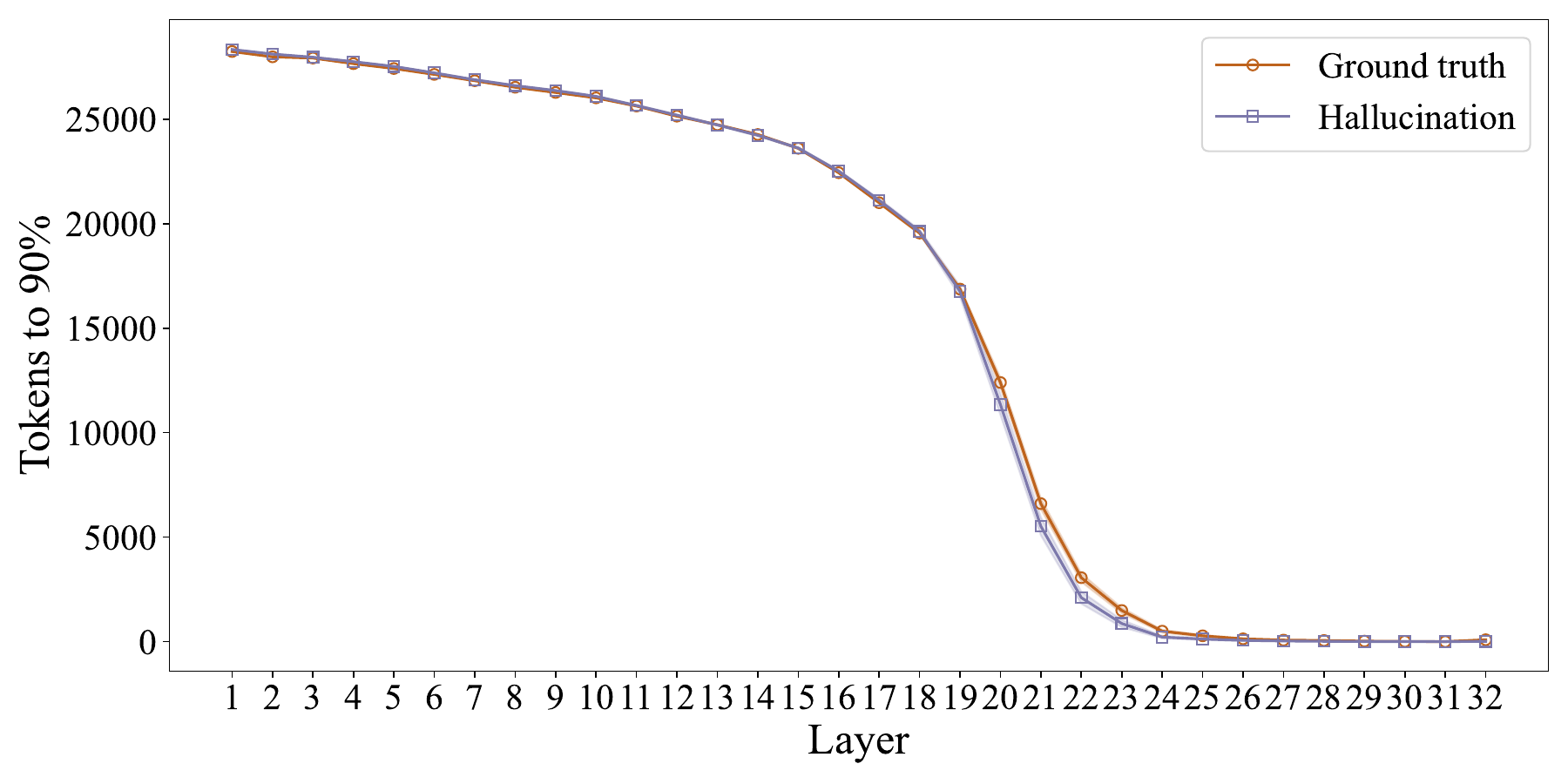}
\caption{The trajectories of the top $90\%$ percentile logits across all layers of MiniGPT-4.}
\label{k_logits_minigpt4}
\end{figure}

\begin{figure}[!htbp]
\centering
\includegraphics[width=0.8\columnwidth]{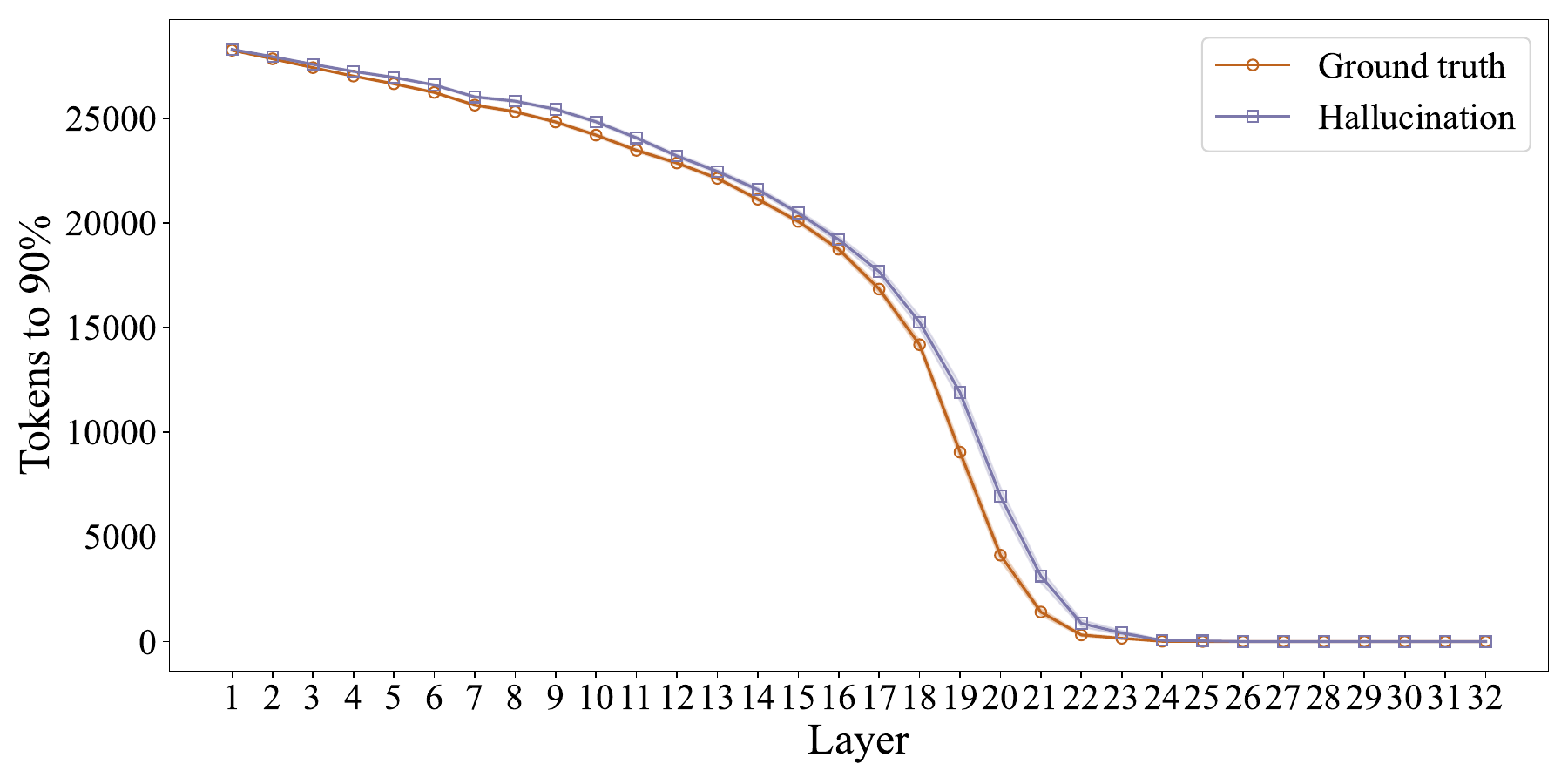}
\caption{The trajectories of the top $90\%$ percentile logits across all layers of Shikra.}
\label{k_logits_shikra}
\end{figure}

We tracked the trajectories of the top $90\%$ percentile logits across all layers of MiniGPT-4 and Shikra (\cref{k_logits_minigpt4} and \cref{k_logits_shikra}). The curves plateau around layer 26, suggesting that content integration and reasoning are effectively completed by the end of the first 25 layers. Accordingly, our hallucination detection and correction mechanisms are concentrated on these initial layers, which confirm the prior hypothesis. 

\begin{table}[!htbp]
  \centering
  \caption{Ablation study of correction layers across different models. The best result is highlighted in bold, and the second-best is underlined.}
  \label{tab4}
  \resizebox{0.45\textwidth}{!}{
  \begin{tabular}{c ccc ccc ccc}
    \hline
    \multirow{2}{*}{Layer} 
      & \multicolumn{3}{c}{LLaVA} 
      & \multicolumn{3}{c}{MiniGPT-4} 
      & \multicolumn{3}{c}{Shikra} \\
    \cline{2-10}
      & C$_S$ & C$_I$ & F1
      & C$_S$ & C$_I$ & F1
      & C$_S$ & C$_I$ & F1 \\
    \hline
    0--10  & 24.0 & \underline{7.0} & 75.1 
           & 18.8 & \textbf{5.6}    & 67.6
           & 47.2 & 15.2             & 70.3 \\
    5--25  & 37.0 & 11.2             & \textbf{77.7}
           & \textbf{12.6} & \textbf{5.6} & 67.3
           & 53.2 & 14.8             & \textbf{74.8} \\
    0--25  & \textbf{20.6} & \textbf{6.2} & \underline{75.3}
           & 21.8 & \underline{6.0}  & \underline{68.1}
           & \textbf{25.2} & \textbf{10.9} & 62.5 \\
    26--31 & 37.0 & 11.2             & 74.7
           & 33.0 & 10.6             & \textbf{71.7}
           & 58.0 & 16.3             & \underline{73.4} \\
    0--31  & \underline{22.2} & 7.8   & 72.5
           & \underline{16.6} & 6.9   & 65.6
           & \underline{28.2} & \underline{11.0} & 64.1 \\
    \hline
  \end{tabular}
  }
\end{table}

\subsection{Full Procedure of D-LEAF}
\label{app:code}

\begin{figure}[!htbp]
\centering
\includegraphics[width=0.8\columnwidth]{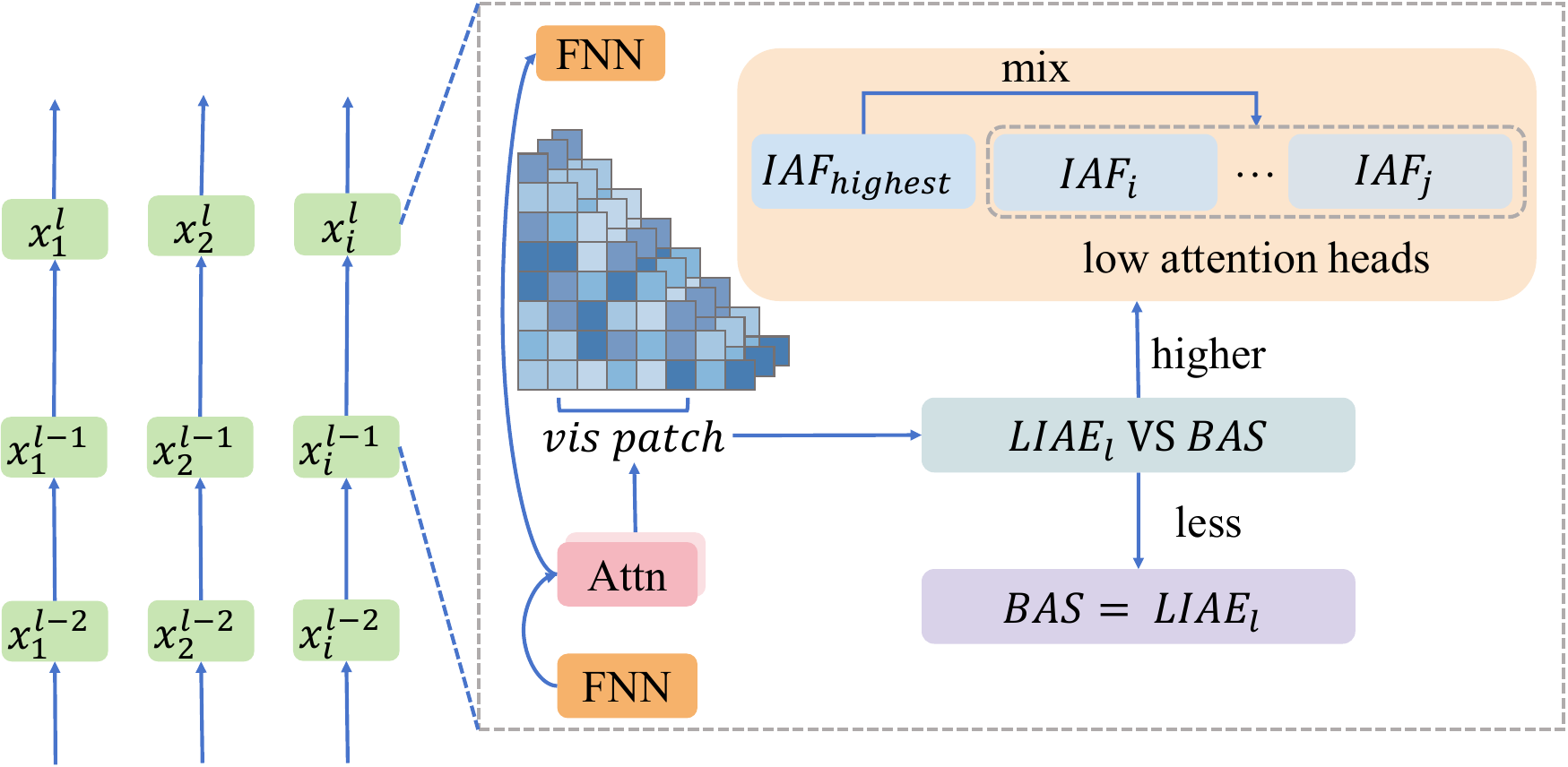}
\caption{Architecture of our D-LEAF.}
\label{alg}
\end{figure}

\cref{alg:algorithm} and \cref{alg} shows the full procedure of D-LEAF.

\begin{algorithm}[!htbp]
\caption{Dynamic Layer-wise Entropy and Attention Fusion}
\label{alg:algorithm}
\textbf{Input}: N input tokens, consisting of both text and vision tokens, each of embedding dimension d, $X \in \mathrm{R}^{(N \times d)}$\\
\textbf{Parameter}: Optional list of parameters\\
\textbf{Output}: The probability distribution of the next token.
\begin{algorithmic}[1] 
\STATE Let $l=0$, $\text{BAS}_{best}=\text{Inf}, A_{tot}=\text{None}$.
\WHILE{$l < L$}
\STATE $Q, K, V = W_QX, W_KX, W_VX$
\FOR{$h \in H$}
\STATE $A_{h} = \frac{Q^{h} (K^{h})^T}{\sqrt{d_k/H}}$
\IF{$A_{tot} = \text{None}$}
\STATE $A_{tot} = A_h$
\ELSE
\STATE $A_{tot} = \text{concat}(A_{tot}, A_h)$
\ENDIF
\ENDFOR
\IF{$l \in (l_s, l_e)$}
\STATE $\text{BAS}_{l} = \text{LIAE}^{l}$
\IF {$\text{BAS}_{best} < \text{BAS}_{l}$}
\STATE $\text{BAS}_{best} = \text{BAS}_{l}$
\ELSE
\STATE $A_{set} = \text{sort}(A_{tot}, \text{key}=\text{IAF}, \text{asc}=\text{True})$
\FOR{$A_h \in A_{set}[0:n]$}
\STATE $A_{h,v} = A_{h}[I_{start}:I_{end}]$
\STATE $A_{best,v} = A_{set}[-1][I_{start}:I_{end}]$
\STATE $A_{h,v} = \gamma A_{best,v} + (1-\gamma) A_{h,v}$
\ENDFOR
\ENDIF
\ENDIF
\STATE $X = \text{FFN}(A_{tot}V)$
\STATE $l = l+1$
\ENDWHILE
\STATE \textbf{return} $X$
\end{algorithmic}
\end{algorithm}

\section{Connection with DPO}
\label{appcntdpo}
In this section, we present the conceptual connection between D-LEAF and DPO. Our analysis is based on the theoretical framework of \cite{uppaal2024model, yang2025nullu} and  where a simple logistic model is adopted to represent the probability of the output token given the continuing prompt. Similar to the main text, the subsequent analysis will be conducted separately for each layer.

Given an input prompt along with its hallucinated and non-hallucinated responses, the corresponding embeddings are denoted as $x$, $y^+$, $y^-$. The loss function optimized in DPO is defined as: 
\begin{equation}
\label{appdpoloss}
\begin{aligned}
\mathcal{L}_{\text{DPO}}(\pi_{\theta}; \pi_{\text{ref}}) 
&= -\,\mathbb{E}_{(x, y^+, y^-) \sim D} \bigg[
    \log \sigma\!\left(
    \beta \log \frac{\pi_{\theta}(y^+|x)}{\pi_{\text{ref}}(y^+|x)}
    \right)  \\
&\quad -\, \beta \log 
    \frac{\pi_{\theta}(y^-|x)}{\pi_{\text{ref}}(y^-|x)}
\bigg]
\end{aligned}
\end{equation}
where, $\pi_{\text{ref}}$ corresponds to the reference (or base) probability model generating output $y$ given $x$, $\pi_{\theta}$ is the new probability model (parametrized by $\theta$), $\sigma$ is the logistic function with $\sigma(z) = (1 + \exp(-z))^{-1}$, and $\beta > 0$ is a hyperparameter. The gradient of the loss $L_\text{DPO}$ with respect to $\theta$ at initialization $\pi_{\theta} = \pi_{\text{ref}}$ equals
\begin{equation}
\label{appdpograd}
\begin{aligned}
\nabla_{\theta} \mathcal{L}_{\text{DPO}}(\pi_{\theta}; \pi_{\text{ref}}) 
\big|_{\pi_{\theta} = \pi_{\text{ref}}} 
= \\ -\,\beta\,\mathbb{E}_{(x, y^+, y^-) \sim D}
\big[
\nabla_{\theta} \log \pi & _{\theta}(y^+|x) - \nabla_{\theta} \log \pi_{\theta}(y^-|x)\big] 
\big|_{\pi_{\theta} = \pi_{\text{ref}}}
\end{aligned}
\end{equation}
For the language decoder of the MLLMs, let $\mathcal{V}$ represent the vocabulary. We begin with a prompt $x \in \mathcal{V}$ and sequentially generate M next-token predictions $y_1, y_2, ..., y_M \in \mathcal{V}$. At each step, the model predicts the token $y_m$based on the input $x_m = (x, y_1, y_2, ..., y_{m-1})$, where $\boldsymbol{x}_m$ denotes the encoded representation of the current prompt. We assume a logistic model that generates each continuation $y_m$ conditioned on $x_m$ defined as follows.
\begin{equation}
\label{appdpopred}
\pi_{\theta}(y_m \mid \boldsymbol{x}_m) 
\equiv 
\pi_{\boldsymbol{W}}(y_m \mid \boldsymbol{x}_m)
= 
Z_{m,\boldsymbol{W}}^{-1}
\exp\!\left(
\boldsymbol{o}_{y_m}^{\top} \boldsymbol{W} \boldsymbol{x}_m
\right)
\end{equation}
Here, $\boldsymbol{o}_{y_m}$ is the classification vector which we use to get
the final word prediction, $\boldsymbol{W}$ is a weight matrix and $Z_{m,\boldsymbol{W}}$ is the normalizing constant:
\begin{equation}
\label{appdpocoeff}
Z_{m,\boldsymbol{W}} = 
\sum_{y \in \mathcal{V}} 
\exp\!\left(
\boldsymbol{o}_{y_m}^{\top} \boldsymbol{W} \boldsymbol{x}_m
\right)
\end{equation}
In \cref{appdpopred}, we assume the classifier is performed with linear transformed module rather than the more common non-linear transformations in the transformer architecture. Based on this model, we then obtain $y = (y_1, y_2, ..., y_M)$ from the prompt $x$ as follows:
\begin{equation}
\label{appdpojoint}
\begin{aligned}
\pi_{\theta}(y \mid x)
&\equiv 
\pi_{\boldsymbol{W}}(y \mid x)
= 
\prod_{m=1}^{M} 
\pi_{\boldsymbol{W}}(y_m \mid \boldsymbol{x}_m)
\\
&= 
Z_{\boldsymbol{W}}^{-1}
\exp\!\left(
\sum_{m=1}^{M}
\boldsymbol{o}_{y_m}^{\top} 
\boldsymbol{W} 
\boldsymbol{x}_m
\right),
\end{aligned}
\end{equation}

where $Z_{\boldsymbol{W}} = \prod_{m=1}^{M} Z_{m,\boldsymbol{W}}$. We denote by $\boldsymbol{x}_m^{\pm}$, $\boldsymbol{x}_m^{\pm}$ and $\boldsymbol{o}_{y_m}^{\pm}$ the positive/negative inputs, the corresponding embedding and classification vector for the positive/negative continuation respectively. Plugging this into \cref{appdpograd}, the first step DPO update has gradient:
\begin{equation}
\label{appdpogradW}
\begin{aligned}
\nabla_{\boldsymbol{W}} \mathcal{L}_{\text{DPO}}(\pi_{\boldsymbol{W}}; \pi_{\text{ref}})
\big|_{\pi_{\boldsymbol{W}} = \pi_{\text{ref}}} \\
= 
-\,\beta\,\mathbb{E}_{(x, y^+, y^-) \sim D}
\Bigg[
\sum_{m=1}^{M}
\Big(
\boldsymbol{o} & _{y_m^{+}} (\boldsymbol{x}_m^{+})^{\top}
-
\boldsymbol{o}_{y_m^{-}} (\boldsymbol{x}_m^{-})^{\top}
\Big)
\Bigg].
\end{aligned}
\end{equation}
Note that the normalization factors $Z_{m,\boldsymbol{W}}$ (and hence $Z_{\boldsymbol{W}}$) are omitted when we take the difference of the gradients of the log-probabilities. 
With $N$ pairs of inputs in $\mathcal{D}$, and we consider the case $M = 1$, the DPO gradient will be an average over all the pairs:

\begin{equation}
\label{appdpogradavg}
\begin{aligned}
\nabla_{\boldsymbol{W}} \mathcal{L}_{\text{DPO}}(\pi_{\boldsymbol{W}}; \pi_{\text{ref}})
\big|_{\pi_{\boldsymbol{W}} = \pi_{\text{ref}}} \\
= 
-\frac{\beta}{N}
\sum_{i=1}^{N}
\Big(
& \boldsymbol{o}_{y_i^{+}} (\boldsymbol{x}_i^{+})^{\top}
-
\boldsymbol{o}_{y_i^{-}} (\boldsymbol{x}_i^{-})^{\top}
\Big),
\end{aligned}
\end{equation}
where the extra index $i$ indicates the $i$-th sample pair. The \cref{appdpogradavg} corresponds to the \cref{thm:dpo_dleaf} in our main paper, which is
\begin{equation}
\label{appdpogradfinal}
\begin{aligned}
\nabla_{\boldsymbol{W}} \mathcal{L}_{\text{DPO}}
&= 
-\frac{\beta}{N}
\sum_{i=1}^{N}
\Big(
\boldsymbol{o}_{y_i^{+}} (\boldsymbol{x}_i^{+})^{\top}
-
\boldsymbol{o}_{y_i^{-}} (\boldsymbol{x}_i^{-})^{\top}
\Big)
\\
= 
-\frac{\beta}{N}
& \sum_{i=1}^{N}
\Big(
\underbrace{
\boldsymbol{o}_{y_i} (\boldsymbol{x}_i^{+} - \boldsymbol{x}_i^{-})^{\top}
}_{\text{feature difference}}
+
\underbrace{
(\boldsymbol{o}_{y_i^{+}} - \boldsymbol{o}_{y_i^{-}}) (\boldsymbol{x}_i^{-})^{\top}
}_{\text{output difference}}
\Big).
\end{aligned}
\end{equation}

\cref{thm:dpo_dleaf} shows that D-LEAF improves the embeddings by adjusting the underperforming attention heads, making them closer to the embeddings produced when the model generates the preferred output. Therefore, both D-LEAF and DPO mitigate hallucinations by modifying the embeddings so that the model’s outputs align more closely with the preferred responses.

\section{Case Study}
\label{app:case}
\begin{figure*}[!htbp]
\centering
\includegraphics[width=\textwidth]{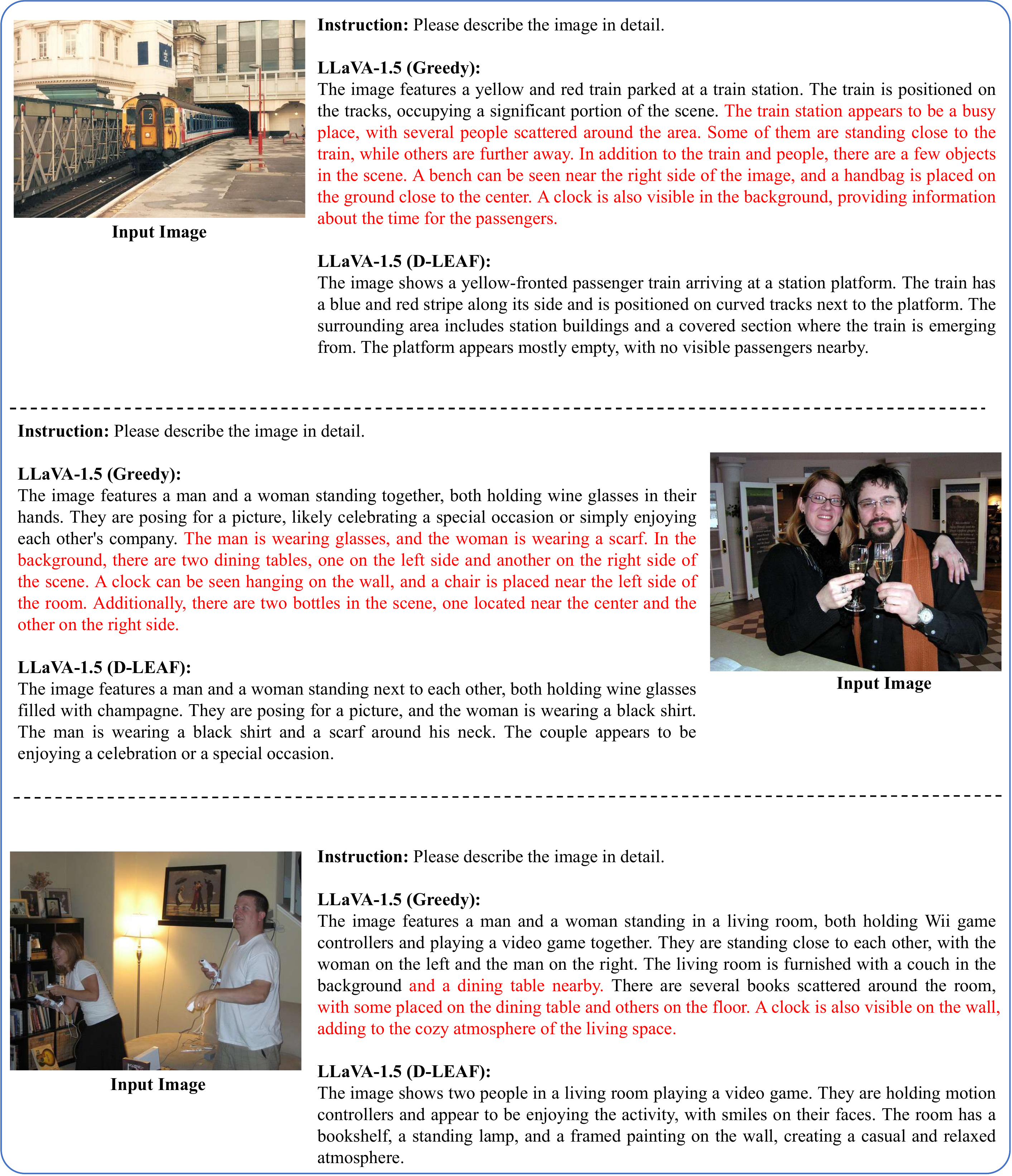}
\caption{D-LEAF’s performance on reducing hallucinations of LLaVA-1.5.}
\label{llava_case}
\end{figure*}

\begin{figure*}[!htbp]
\centering
\includegraphics[width=\textwidth]{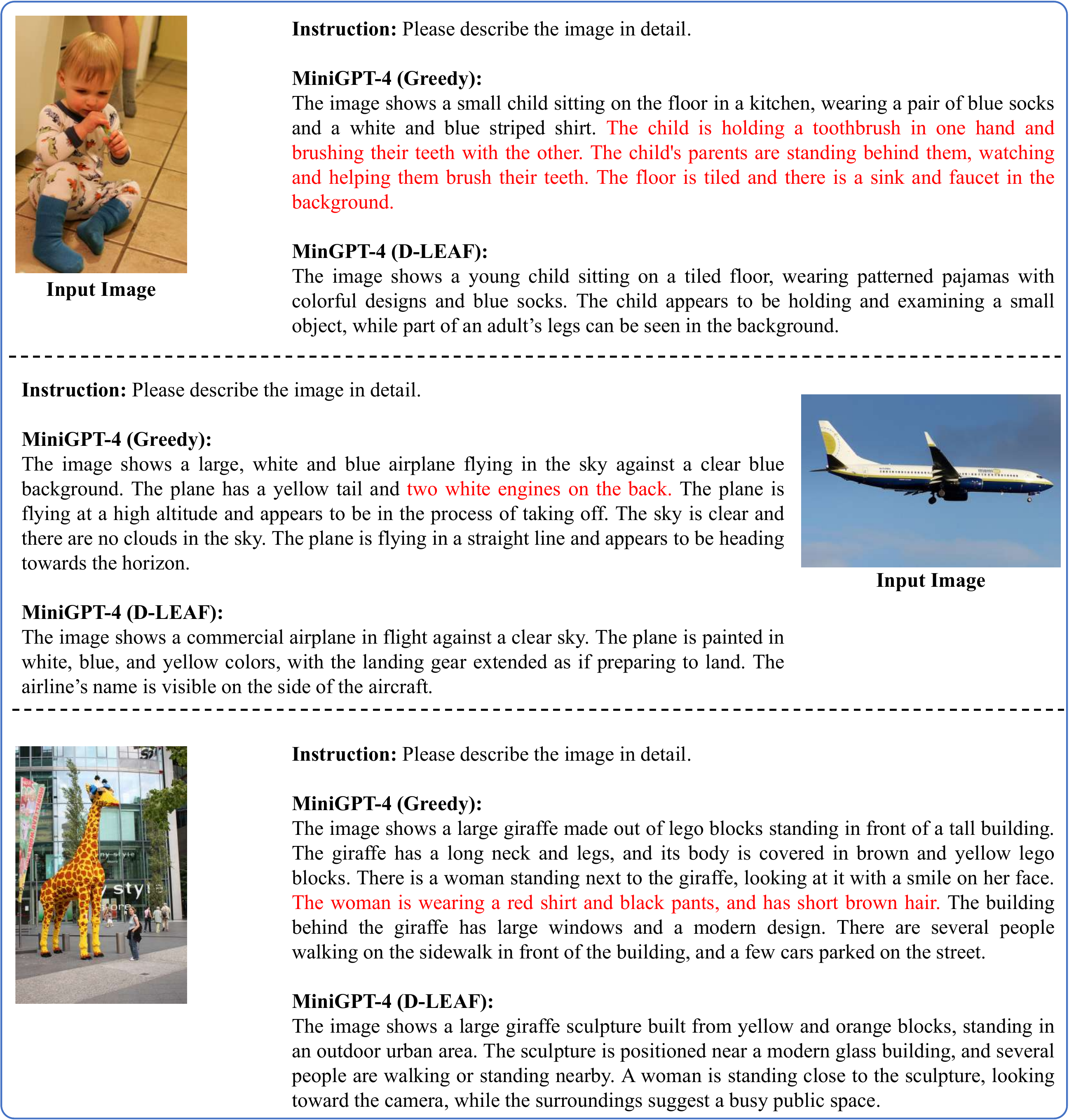}
\caption{D-LEAF’s performance on reducing hallucinations of MiniGPT4.}
\label{minigpt4_case}
\end{figure*}

\begin{figure*}[!htbp]
\centering
\includegraphics[width=\textwidth]{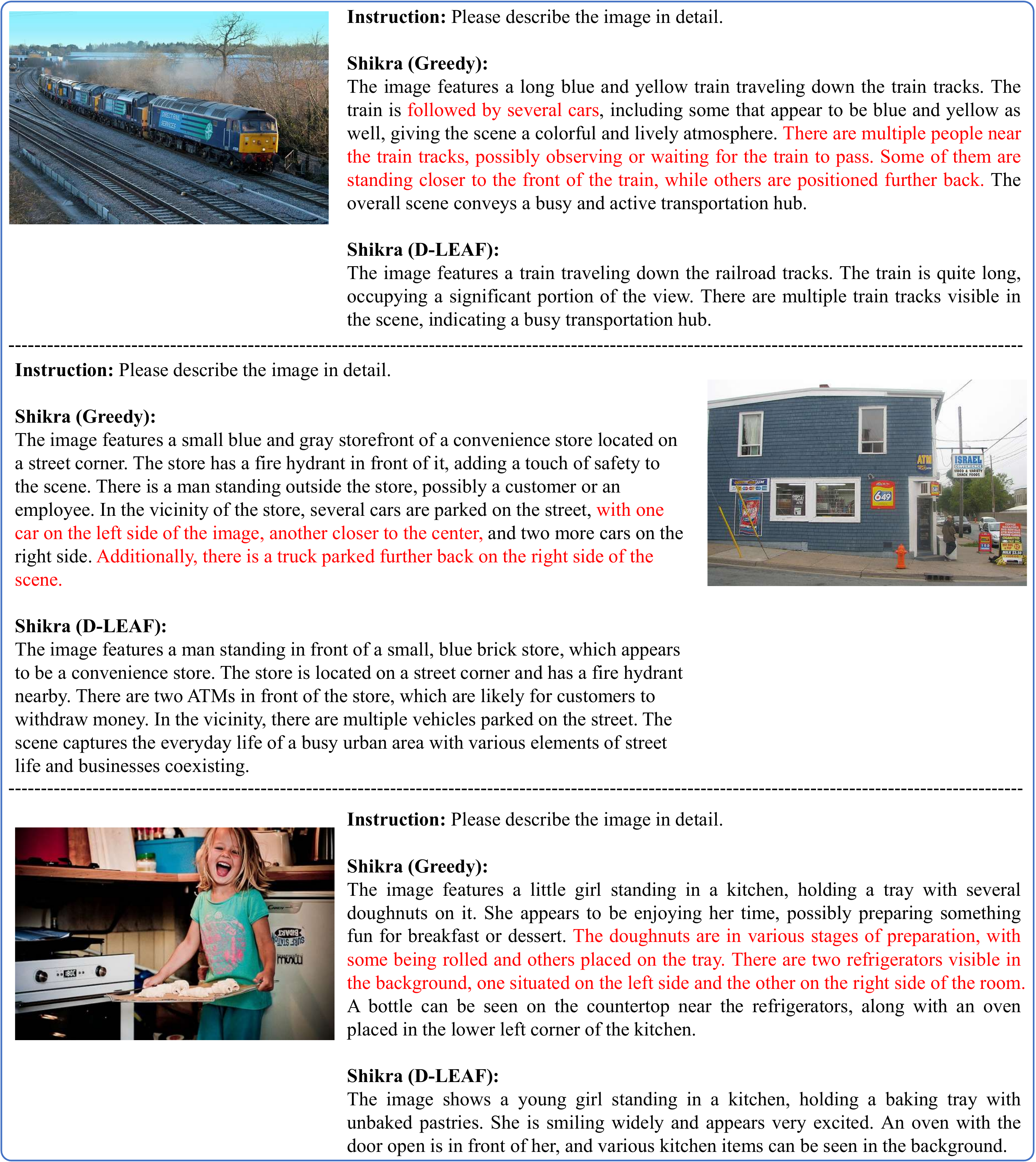}
\caption{D-LEAF’s performance on reducing hallucinations of Shikra.}
\label{shikra_case}
\end{figure*}

Additional case studies across diverse MLLMs are provided to demonstrate the effectiveness of D-LEAF. Specifically, \cref{llava_case}, \cref{minigpt4_case} and \cref{shikra_case} respectively present further instances of hallucination corrections by our method for LLaVA-1.5, MiniGPT-4 and Shikra.

\end{document}